\pdfoutput=1
\documentclass[preprint,12pt]{elsarticle}

\usepackage{graphicx}
\usepackage{wrapfig}
\usepackage{lscape}
\usepackage[algochapter,linesnumbered,ruled,lined,boxed]{algorithm2e}
\usepackage{float}

\usepackage{amssymb}

\usepackage{multirow}
\usepackage{longtable}
\usepackage{tabu}
\usepackage{diagbox}
\usepackage[algochapter,linesnumbered,ruled,lined,boxed]{algorithm2e}

\usepackage{color}

\usepackage{booktabs}
\usepackage{caption}
\usepackage{subfigure}
\usepackage{verbatim}
\usepackage{mathtools}
\newsavebox\CBox

\usepackage{url}
 
\journal{}

\begin{document}

\begin{frontmatter}


\title{Supervised Contrastive Learning with Tree-Structured Parzen Estimator 
 Bayesian Optimization for Imbalanced Tabular Data }




\author{Shuting Tao}
\ead{12121105@zju.edu.cn}
\author{Peng Peng\corref{mycorrespondingauthor}}
\ead{pengp17@mails.tsinghua.edu.cn}
\author{Qi Li}
\ead{liqi177@zju.edu.cn}
\author{Hongwei Wang\corref{mycorrespondingauthor}}
\ead{hongweiwang@intl.zju.edu.cn}
\cortext[mycorrespondingauthor]{Corresponding author}
\address{Zhejiang University - University of Illinois at Urbana-Champaign Institute, Zhejiang Univeristy, 718 East Haizhou Road, Haining, 314400, China}

\begin{abstract}
Class imbalance has a detrimental effect on the predictive performance of most supervised learning algorithms as the imbalanced distribution can lead to a bias preferring the majority class. To solve this problem, we propose a Supervised Contrastive Learning (SCL) method with Tree-structured Parzen Estimator (TPE) technique for imbalanced tabular datasets. Contrastive learning (CL) can extract the information hidden in data even without labels and has shown some potential for imbalanced learning tasks. SCL further considers the label information based on CL, which also addresses the insufficient data augmentation techniques of tabular data. Therefore, in this work, we propose to use SCL to learn a discriminative representation of imbalanced tabular data. Additionally, the hyper-parameter temperature $\tau$ of SCL has a decisive influence on the performance and is difficult to tune. We introduce TPE, a well-known Bayesian optimization technique, to automatically select the best $\tau$. Experiments are conducted on both binary and multi-class imbalanced tabular datasets. As shown in the results obtained, TPE outperforms three other hyper-parameter optimization (HPO) methods such as grid search, random search, and genetic algorithm. More importantly, the proposed SCL-TPE method achieves much-improved performance compared with the state-of-the-art methods.
\end{abstract}

\end{frontmatter}

\section{Introduction}
\label{sec:intro}

With excellent performance on uniformly distributed data, supervised learning has become the most popular method for data classification. However, uneven distribution of data, i.e., class imbalance, is very common in most datasets collected from real-world scenarios, which inevitably undermines the effectiveness of supervised algorithm. This class imbalance makes it intractable for the supervised models to represent the distribution characteristics of skewed data correctly, and thus results in very low prediction accuracy for the minority classes. A well-known example is the mammography dataset ~\cite{he2009learning}, in which positive samples only account for 2.3$\%$ of the total samples. While the prediction accuracy of the positive class is crucial in this case, the traditional supervised learning-based classifiers tend to predict that all samples are negative. 

Improving classification accuracy of both the majority and minority classes has become a great challenge. To address this challenge, considerable solutions have been put forward and they can be broadly divided into three categories: data preprocessing, \cite{chawla2002smote,mani2003knn,han2005borderline,he2008adasyn}, feature learning \cite{zhang2022intelligent,ng2016dual}, and classifier design\cite{elkan2001foundations,shrivastava2016training,lin2017focal}. For the solutions based on data preprocessing, scholars attempt to rebalance data distribution through data sampling. In terms of classifier design, there are two kinds of methods. Specifically, the algorithm-level methods modify algorithms to increase the low accuracy of the minority class - the most popular one is cost-sensitive learning (CSL) that uses a weighted cost for different classes. Another kind is the model-level methods which combine the classification results from multiple base models like ensemble learning.

However, the methods mentioned above have some inherent drawbacks. For example, the data-level methods may result in losing useful information \cite{liu2008exploratory} or overestimating the minority data \cite{liu2021modified}. For CSL, it is difficult to set the value of misclassification cost which in most cases is unknown from the data and cannot be given by experts \cite{haixiang2017learning}. These issues have motivated researchers to develop strategies based on feature learning, and the existing methods consider using autoencoders to learn imbalanced data features \cite{ng2016dual}. In this paper, we propose to use supervised contrastive learning (SCL) \cite{khosla2020supervised} to extract features from imbalanced tabular datasets. 

Contrastive learning (CL), a kind of self-supervised learning (SSL) \cite{liu2021self}, has shown the ability to represent the hidden features that are not conditioned on data labels in the image domain \cite{chen2020simple, VikashSehwag2021SSDAU}. CL aims to group an anchor and a ``positive" sample together in the embedding space, and diverse the anchor far from ``negative" samples. Here ``positive" sample refers to data augmented from the anchor, while ``negative" samples are randomly chosen from small batches. It is noted that the success achieved by CL in feature learning of images is closely related to data augmentation techniques, such as rotation \cite{gidaris2018unsupervised}, colorization \cite{zhang2016colorful}, and jigsaw puzzle solving \cite{noroozi2016unsupervised}. And most of these techniques are not applicable to general tabular data because they heavily rely on the unique structure of the domain datasets. This brings a great difficulty when using CL for imbalanced tabular learning.

In this work, we fill this gap by adopting SCL to learn the representation of imbalanced tabular data. SCL considers many positives per anchor rather than using only a single positive. These positives are selected from samples belonging to the same class as the anchor, instead of from data augmentations of the anchor. Embeddings of the same class are pulled closer together than those from other classes. The utilization of label information can alleviate the lack of data augmentation strategy of tabular data. So the success of contrastive loss in the image domain loss can be extended to the tabular domain using some domain-independent augmentations such as gaussian blur.

Furthermore, SCL requires fixing the hyper-parameter temperature $\tau$ before model training, which is a crucial hyper-parameter to control the strength of penalties on negative samples. As evident in \cite{FengWang2021UnderstandingTB}, a good choice of $\tau$ can significantly improve the quality of feature representation. That is to say, a good selection of $\tau$ can make the SCL achieve better performance in imbalanced learning. However, hyper-parameter tuning is often challenging and time-consuming. And among the current studies, little consideration has been provided to details of the hyper-parameter tuning of $\tau$. In this paper, we demonstrate that the setting of the $\tau$ substantially influences SCL's performance. We further propose to develop a flexible approach that enables hyper-parameter optimization (HPO) to be conducted as an automatic process. 

Classic HPO methods include grid search (GS), random search (RS), genetic algorithm (GA), and Bayesian optimization (BO). GS defines a search space as a grid of hyper-parameter values and assesses every position in the grid. RS defines a search space as a bounded domain of hyper-parameter values and randomly samples points in that domain. GA is based on the concepts of biological evolution, which considers a set of possible candidate solutions that evolves and gives a better result \cite{whitley1994genetic}. Compared with uninformed search GS and RS, BO considers the previously explored information in each step, which reduces the search space and improves the search efficiency. Compared with GA, BO requires fewer computational resources. GA needs to train the model on multiple hyper-parameters to go from one generation to the next. In contrast, BO trains a single model and updates the posterior information, shortening training time and not requiring many computational resources. In general, BO has two implementations: Gaussian Process (GP) and Tree-structured Parzen estimator (TPE) \cite{bergstra2011algorithms}. TPE has been proven superior to GP since TPE's modeling of previously explored observations is more accurate than GP's \cite{bergstra2011algorithms}. Therefore, we choose TPE to select the SCL model's best $\tau$ in our work. More empirical work is shown in Section \ref{sec:experiment} to confirm our choice. More specifically, the main contributions of this paper are listed below:

\begin{itemize}
    \item SCL is proposed to learn an embedding space in which samples of the same class pairs stay close to each other while samples belonging to different classes are far apart. For imbalanced tabular datasets, we believe that SCL will outperform traditional supervised methods - the reason for this is that, in addition to employing the label information, SCL better captures data features by learning the intrinsic properties from the data itself based on contrastive loss. Therefore, SCL will not suffer from a significant performance drop due to the ``label bias" caused by imbalanced data.
  
    \item  TPE  is first used to select the best hyper-parameter temperature $\tau$ for SCL automatically. In this paper, we demonstrate that $\tau$ is critical to SCL's performance, and TPE is proven to produce better results than other algorithms for hyper-parameter optimization.
    
    \item Extensive experiments are conducted to demonstrate the effectiveness of our method. We compare SCL-TPE's performance with ten competitive data sampling methods on fifteen imbalanced tabular datasets covering binary and multi-class tasks. We further carry out an ablation study to analyze the performance improvement of each component in SCL-TPE. 
\end{itemize}

The rest of this article is organized as follows. In Section~\ref{sec:background}, a brief review of previous research targeting the imbalanced learning problem is described. We also describe SCL and TPE Bayesian optimization as the theoretical foundation of the proposed SCL-TPE method. Section~\ref{sec:method} presents the proposed method in detail. Section~\ref{sec:experiment} evaluates the proposed method by conducting experiments on some highly imbalanced datasets. Finally, the main conclusions of this work are drawn and discussed in Section~\ref{sec:conclusion}.

\section{Related work and background theory}
\label{sec:background}
\begin{figure}[!ht]
\centering
\includegraphics[width=0.8\textwidth]{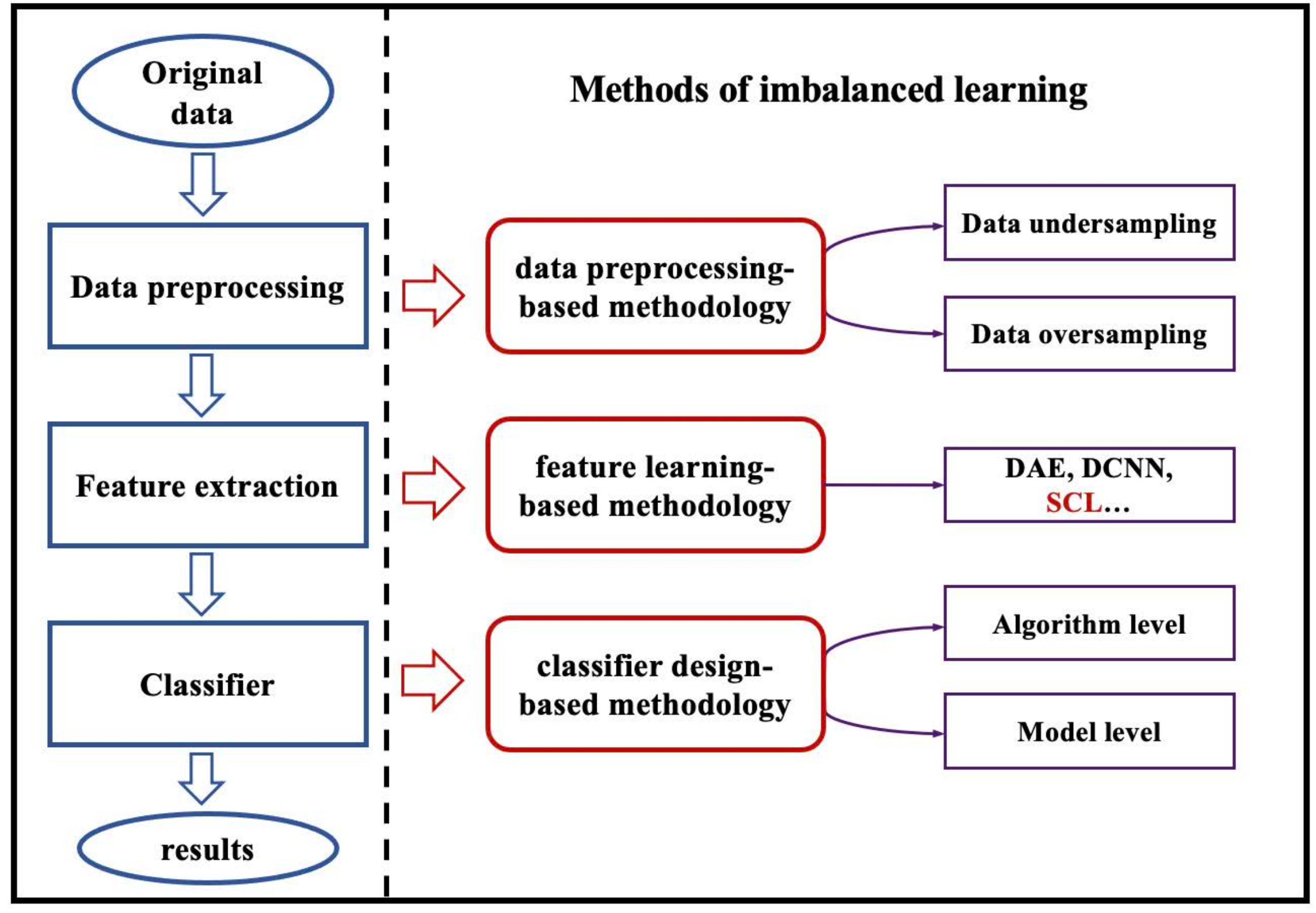}
\caption{The three methods of imbalanced learning.} 
\label{methods}
\end{figure}
\subsection{Methods for imbalanced learning}
This subsection briefly reviews the related work on imbalanced learning methods. As shown in Fig. \ref{methods}, countermeasures to mitigating class imbalance issues can be divided into three categories: methods based on data preprocessing, methods based on feature learning, and methods based on classifier design.
\subsubsection{Methods based on data preprocessing}
Data-level methods conduct preprocessing on imbalanced datasets by modifying data distribution through sampling. These methods can be further divided into two sub-groups: undersampling and oversampling.

Undersampling methods rebalance data distribution by removing instances of the majority class. A non-heuristic approach is random undersampling (RUS) which randomly removes some majority class examples. However, such a manner may lead to the loss of important information in those removed examples. To alleviate this problem, more advanced undersampling techniques have been developed, such as Edited Nearest Neighbor (ENN) \cite{wilson1972asymptotic}, Condensed Nearest Neighbor (CNN) \cite{hart1968condensed} , Tomek Link \cite{ITomek1976TwoMO}, One-Sided Selection (OSS) \cite{kubat1997addressing}. ENN investigates k-nearest neighbors for each instance of the majority class. If most of the k neighbors belong to a different class, the instance will be removed. CNN is achieved by enumerating the dataset and adding them to the ‘group’ only if they cannot be classified correctly by the current contents. If there are two instances of different classes whose nearest neighbors are each other, they form a Tomek Link. In the Tomek Link approach, all instances in Tomek links that belong to the majority class are removed. OSS uses the Tomek Link method on top of CNN to eliminate noisy data and rebalance the data distribution.

Oversampling methods aim at increasing the size of the minority class by generating artificial minority instances. The simplest technique is random oversampling (ROS), which randomly replicates the minority class examples to rebalance the original imbalanced datasets. Due to generating duplicate data, ROS is prone to overfitting. In this regard, \cite{chawla2002smote} introduces Synthetic Minority Oversampling (SMOTE). SMOTE generates synthetic instances by interpolating between a minority class instance and its k nearest minority class neighbors. However, the main limitation of SMOTE is that each minority class example generates the same number of artificial samples without considering neighboring examples can come from different classes, which may result in overlap between classes. To accommodate this scenario, Borderline-SMOTE (BSMOTE) \cite{han2005borderline} and Adaptive Synthetic Sampling (ADASYN) \cite{he2008adasyn} algorithms have been designed. BSMOTE only creates new data for minority instances close to the border. ADASYN, on the other hand, leverages data distribution to determine the number of samples to be synthesized for each minority sample. Generative models such as generative adversarial networks (GAN) can also be used to generate synthetic minority class samples for data oversampling. This network synthesis-based approach is more complex than traditional techniques, but the generated samples are more diverse.

\subsubsection{Methods based on feature learning}
Feature-learning-based strategy attempts to preserve the key features of data to increase the discrimination between the minority class and the majority class. Using neural networks for feature extraction targeting imbalanced datasets has led to many in-depth studies. In these achievements, deep convolutional neural networks (DCNN) and deep autoencoders (DAE) are employed as the basic models. For example, \cite{ng2016dual} proposed the Dual Autoencoding Features (DAF), a feature learning method based on the stacked auto-encoder, to learn features with better classification capabilities of the minority and the majority classes.   

In this work, we use the SCL method to learn the features of imbalanced data since SCL can utilize the rich implicit information from data as well as the information provided by labels. We will introduce the technical details in Section~\ref{subsec:SCL}.

\subsubsection{Methods based on classifier design}
Classifier-design-based methodology involves algorithm-level methods and model-level methods. Algorithm-level methods assign different weights for the majority and minority classes, thus easing the optimization difficulty under imbalanced data \cite{cui2019class,tang2020long}. Model-level methods focus on constructing models that are less sensitive to imbalanced data. Among them, models built by ensemble approaches have become popular in imbalanced learning due to their better performance than a single learner \cite{JafarTanha2020BoostingMF}, and the pure ensemble method is usually combined with algorithm-level method or data-preprocess-based strategy. For instance, EasyEnsemble is an ensemble solution embedded with RUS.

\subsection{Supervised contrastive learning }
\label{subsec:SCL}
Many researchers have employed CL methods in previous studies to learn data representations by attracting positive pairs and pushing apart negative pairs. To optimize for this property, self-supervised contrastive learning has been proposed, which contrasts a single positive sample for each anchor against a set of negatives consisting of the entire remainder of the batch in the embedding space. The positive sample is an augmented version of the anchor. 

Suppose there is a batch of $N$ samples with their labels, $\left\{\boldsymbol{x}_{k}, \boldsymbol{y}_{k}\right\}_{k=1 \ldots N}$. For each input sample $x$, two random augmentations are generated, so the augmented batch used for training comprises $2 N$ pairs, $\left\{\tilde{\boldsymbol{x}}_{\ell}, \tilde{\boldsymbol{y}}_{\ell}\right\}_{\ell=1 \ldots 2 N}$, among which $\tilde{\boldsymbol{x}}_{2 k}$ and $\tilde{\boldsymbol{x}}_{2 k-1}$ are augmentations 
of $\boldsymbol{x}_{k} {(k=1 \ldots N)}$ and 
$\tilde{\boldsymbol{y}}_{2 k-1}=\tilde{\boldsymbol{y}}_{2 k}=\boldsymbol{y}_{k}$. 
In the augmented batch, we assume $i \in I \equiv\{1 \ldots 2 N\}$ be the index of an arbitrary augmented sample and $j(i)$ be the index of 
the other augmented sample originating from the same source sample. The self-supervised contrastive loss \cite{chen2020simple} is given by:

\begin{equation}
\label{equ:SSCL}
\mathcal{L}^{s e l f}=\sum_{i \in I} \mathcal{L}_{i}^{s e l f}=-\sum_{i \in I} \log \frac{\exp \left(\boldsymbol{z}_{i} \cdot \boldsymbol{z}_{j(i)} / \tau\right)}{\sum_{a \in A(i)} \exp \left(\boldsymbol{z}_{i} \cdot \boldsymbol{z}_{a} / \tau\right)}
\end{equation}
 In Eq. (\ref{equ:SSCL}), $\boldsymbol{z}_{\ell}=\operatorname{Enc}\left(\tilde{\boldsymbol{x}}_{\ell}\right) $, $\operatorname{Enc}\left(\cdot\right)$ is the representation vector of  $\tilde{ \boldsymbol{x}}_{\ell}$, temperature $\tau \in \mathcal{R}^{+}$ is a scalar  parameter, and $A(i) \equiv I \backslash\{i\}$. The index $i$ denotes the anchor, index $j(i)$ denotes the positive, and the other $2(N-1)$ indices $(k \in A(i) \backslash\{j(i)\})$ denote the negatives. Note that for each anchor $i$, the denominator has a total of $2N-1$ terms consisting of one positive sample and $2N-2$ negative samples.

In this study, we propose to utilize SCL, a generalization of self-supervised contrastive loss \cite{khosla2020supervised}. SCL leverages the label information and contrasts the set of all samples from the same class as positives against the negatives from other classes. The formulation of supervised contrastive loss is:

\begin{equation}
\label{equ:SCL}
\mathcal{L}^{\text {sup }}=\sum_{i \in I} \frac{-1}{|P(i)|} \sum_{p \in P(i)} \log \frac{\exp \left(\boldsymbol{z}_{i} \cdot \boldsymbol{z}_{p} / \tau\right)}{\sum_{a \in A(i)} \exp \left(\boldsymbol{z}_{i} \cdot \boldsymbol{z}_{a} / \tau\right)} 
\end{equation}
where $P(i) \equiv\left\{p \in A(i): \tilde{\boldsymbol{y}}_{p}=\tilde{\boldsymbol{y}}_{i}\right\}$ is the set of indices of all positives in the augmented batch distinct from $i$, and $|P(i)|$ is its cardinality. 

\subsection{Tree-structured parzen estimator Bayesian optimization}
It is well recognized that networks are easy to apply but difficult to train. The hyper-parameter tuning can be regarded as a “black art” requiring human experience, trial and error methods, and sometimes even violent search. Therefore, HPO emerges for three purposes: 1. Reduce labor costs; 2. Improve the performance of the model; 3. Help to find more reproducible parameter sets. Among the four HPO approaches mentioned in Section \ref{sec:intro}, BO is an informed search, which uses the performance of the previously searched parameters to speculate the best next step, thus reducing the search space and significantly improving the search efficiency. 

Assume $x$ is the hyper-parameter we want to optimize, and our goal is to minimize the objective function $y = f(x)$. BO uses an iterative method called Sequential Model-Based Global Optimization (SMBO). The algorithm first builds a probability model $M$ for the function we want to optimize. In each subsequent iteration $t$, the algorithm selects the local optimal hyper-parameter $x_{t}$ based on the current model $M_{t-1}$ according to the acquisition function. The acquisition function here defines a balance between exploring new areas in the objective space and exploiting areas already known to obtain the favorable $x$. There are many forms of acquisition function, e.g., the probability of improvement, entropy search, and expected improvement (EI) \cite{jones2001taxonomy}. This paper chooses EI as the criterion due to its excellent and intuitive performance \cite{snoek2012practical}, which is defined as: 

\begin{equation}
\label{equ:EI}
E I_{y^{*}}(x) =\int_{-\infty}^{+\infty} m a x\left(y^{*}-y, 0\right) p(y \mid x) d y .\\
\end{equation}
$y^{*}$ here is a threshold. When ${x}$ is given, EI is the expectation that $f(x)$ will exceed (negatively)  $y^{*}$.  The hyper-parameter $x_{t}$ with the greatest EI is the local optimal hyper-parameter. After finding $x_{t}$, $f(x_{t})$ is evaluated. Then SMBO stores the $x_{t}$ and $f(x_{t})$ into the search history, and fit a new model $M_{t}$ based on the updated history. When the loop ends, SMBO outputs the global optimal hyper-parameter.

Under the framework of SMBO, TPE is proposed \cite{bergstra2011algorithms}. Instead of modeling $M$ as $p(y \mid x)$ directly, TPE uses Bayes' theorem to decompose the  $p(y \mid x)$ into $p(x \mid y)$ and $p(y)$.
  
\begin{equation}
\label{equ:bayes}
p(y \mid x) =\frac{p(x \mid y) p(y)}{p(x)}
\end{equation}

\begin{equation}
\label{equ:threshold}
p(x \mid y) =\left\{\begin{array}{l}
l(x), \quad y<y^{*} \\
g(x), \quad y>y^{*}
\end{array}\right.\\
\end{equation}

As can be seen in Eq. (\ref{equ:threshold}), TPE constructs different $p(x \mid y)$ on different sides of the threshold, where $l(x)$ is the density formed by using the observations ${x}$ such that corresponding loss $f(x)$ is less than $y^{*}$ and $g(x)$ consists of the remaining observations. And the algorithm will set $y^{*}$ to be some quantile $\gamma$ of the observed y values, so that $p(y < y^{*}) = \gamma$.

Combining  Eqs. (\ref{equ:EI}), (\ref{equ:bayes}), (\ref{equ:threshold}), optimization of EI in the TPE algorithm is concluded to be:
\begin{equation}
E I_{y^{*}}(x)=\frac{\gamma y^{*} \ell(x)-\ell(x) \int_{-\infty}^{y^{*}} p(y) d y}{\gamma \ell(x)+(1-\gamma) g(x)} \propto\left(\gamma+\frac{g(x)}{\ell(x)}(1-\gamma)\right)^{-1}
\end{equation}

\section{Methodology}
\label{sec:method}
\subsection{The training of SCL classification model}
\begin{figure}[!ht]
\centering
\includegraphics[width=1.1\textwidth]{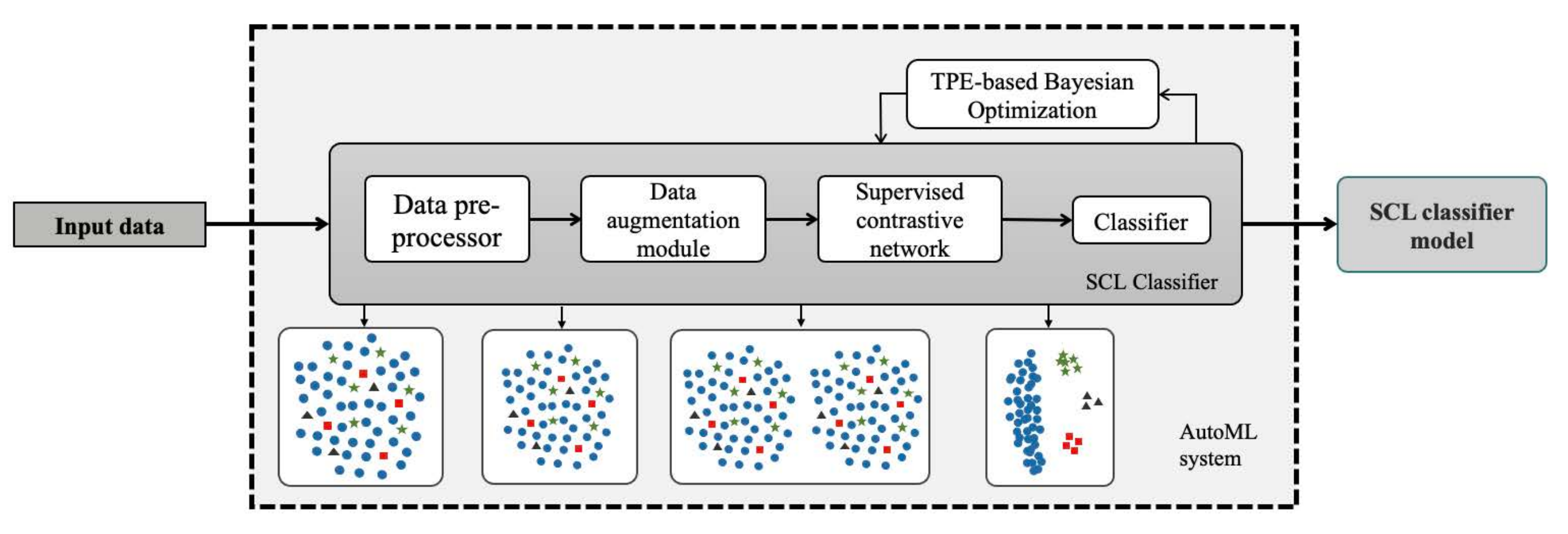}
\caption{The general framework of the SCL-TPE model.} 
\label{framework}
\end{figure}

In this paper, we propose an SCL-TPE for imbalanced tabular datasets. From Fig. \ref{framework}, we can see that the training procedure of the SCL classification model mainly consists of three steps. First, preprocess the original imbalanced dataset. For a multi-feature dataset, we normalize each data feature separately. Second, learn discriminative representations. In this representation learning stage, a data augmentation module is applied to transform each data sample randomly into two correlated instances. Common tabular data transformations, such as gaussian blur are used. If the two augmented instances are from the same sample, they are regarded as positive pair. They are regarded as negative pair if transformed from different instances. Then we train the contrastive network with the augmented dataset, and the supervised contrastive loss helps us obtain a better feature embedding. Third, train the classifier. After acquiring the embedding of each class from the contrastive network, a three-layer softmax classifier will be trained for classification.

\subsection{Determination of temperature $\tau$}
In SCL, temperature $\tau \in \mathcal{R}^{+}$ is a scalar hyper-parameter that significantly impacts the model's performance. Different values of $\tau$ can considerably vary in results. In this work, we use TPE to select the best $\tau$ by minimizing the objective function $y = f(\tau)$. $y$ here represents the negative of the model's area under the receiver operating characteristic curve(AUC), which is an overall metric of the model's performance. We will introduce AUC in Section~\ref{sec:metric}. The smaller value of $y$ means the higher AUC and the better performance of the model. We show the detailed process for determining $\tau$ with TPE in Section \ref{procedure}.

\subsection{The whole procedure of SCL-TPE}
\label{procedure}
The whole process of the proposed method is shown in Algorithm \ref{algorithm}. First, we build a probability distribution model $M_{0}$ and history set $\mathcal{H}$. Then SCL-TPE iterates as follows: in iteration $t$, we first choose the value of $\tau$ according to $M_{t-1}$ and $\mathcal{H}$. The data augmentation module is used to transform the instances. Then we construct the feature extractor and implement the supervised contrastive loss to update the parameters of the extractor network with the selected $\tau$. Subsequently, we train the softmax classifier and apply the trained classifier to test datasets to get the AUC. TPE adds the negative of AUC and the value of $\tau$ to the history set $\mathcal{H}$, and fits a new model $M_{t}$ according to the updated $\mathcal{H}$.

{\footnotesize
\begin{algorithm}
\renewcommand{\thealgocf}{1}
\caption{SCL-TPE for imbalanced datasets}\label{algorithm}
\KwIn{imbalanced training set $(x,y)$, imbalanced test set $(x_{t},y_{t})$, maximum iterations $T$, maximum epochs $n$, batch size $B$, learning rate $\eta$}
\KwOut{Encoder network with learned parameters $ \theta_{1}^{*}$, Classifier network with learned parameters $ \theta_{2}^{*}$, the best hyper-parameter $\tau$}
Initialize $M_{0}$, $\mathcal{H} \leftarrow \emptyset$\;
\For(\tcp*[f]{number of iterations}){$t=1;t\leqslant T$}{
TPE chooses $\tau$ depends on $M_{t-1}$ and $\mathcal{H}$\;
Net, $\theta_{1} \leftarrow$ construct\_ContrastiveNet()\;
\For(\tcp*[f]{number of epochs}){$i=1;i\leqslant n$}
{
 \For(\tcp*[f]{number of batches}){$b=1;b\leqslant B $}
{
 $x_{b},y_{b} \leftarrow$sampling $(x,y)$\; 
 $\tilde{\boldsymbol{x}}_{b}, \tilde{\boldsymbol{y}}_{b} \leftarrow$ data$\_$augmentation $\left (x_{b},y_{b}  \right )$\;
 $\tilde{\boldsymbol{z}}_{b} \leftarrow$forward$(\tilde{\boldsymbol{x}}_{b},$Net$,  \theta_{1},\tau)$\;
 $grad_{  \theta_{1}} \leftarrow$ backward$(\tilde{\boldsymbol{x}}_{b},\tilde{\boldsymbol{z}}_{b},\mathcal{L}^{sup}_{b},$Net$,  \theta_{1})$\;
 $  \theta_{1}^{*} \leftarrow$ update\_NetParams$($Net$, \theta_{1},grad_{ \theta_{1}},\eta)$\;
 $ \theta_{1} \leftarrow  \theta_{1}^{*}$ \;
}
}
Classifier $\leftarrow$ construct\_softmaxclassifier$()$\;
\For(\tcp*[f]{number of epochs}){$i=1;i\leqslant n$}
{
 \For(\tcp*[f]{number of batches}){$b=1;b\leqslant B $}
{
  $x_{b},y_{b} \leftarrow$sampling $(x,y)$\; 
  $\boldsymbol{z}_{b} \leftarrow$ Net $(x_{b})$\;
  $\boldsymbol{O}_{b} \leftarrow$forward$(\boldsymbol{z}_{b},$Classifier$,  \theta_{2})$\;
 $grad_{  \theta_{2}} \leftarrow$ backward$(y_{b},\boldsymbol{O}_{b},\mathcal{L}^{cross}_{b},$Classifier$, \theta_{2})$\;
 $  \theta_{2}^{*} \leftarrow$ update\_ClassifierParams$($Classifier$, \theta_{2},grad_{ \theta_{2}},\eta)$\;
 $ \theta_{2} \leftarrow  \theta_{2}^{*}$ \;
}
}
AUC $\leftarrow $Evaluate(Classifier$(x_{t}),y_{t})$)\;
$\mathcal{H} \leftarrow \mathcal{H} \cup(\tau, -1\times$AUC$))$\;
$\text { Fit a new model } M_{t} \text { to } \mathcal{H} $\;
}
\end{algorithm}
}

\section{Experiment}
\label{sec:experiment}
\subsection{Experimental design}
In this section, three sub-experiments are designed. The first one investigates the effectiveness of TPE. The second one explores whether the proposed SCL-TPE method can outperform the other state-of-the-art algorithms mentioned in Section \ref{sec:background}. The third one conducts an ablation study of SCL, CL-TPE, and SCL-TPE with the same structure.

\subsubsection{Datasets and baselines}
\label{sec:datasets and baselines}
In our experiment, SCL-TPE is evaluated on eight binary and seven multi-class imbalanced datasets collected from the KEEL \cite{alcala2011keel} and UCI \cite{asuncion2007uci} repositories. The detailed information of these fifteen datasets is given in Table \ref{data}.

As for baselines, we introduce three commonly used HPO algorithms in the first sub-experiment: random search, grid search, and genetic algorithm. TPE and random search are performed using the hyperopt package for Python. In the second sub-experiment, we compare the results of the proposed method against ten data sampling techniques, including random sampling, random undersampling, ENN, CNN, OSS, random oversampling, SMOTE, ADASYN, BSMOTE, and GAN. Each sampling method is tested with five classification algorithms, including multilayer perceptron (MLP) with the same structure as the proposed model, support vector machine (SVM), K nearest neighbors (KNN), decision tree classifier (DTC), random forest classifier (RFC). The package imbalanced-learn \cite{JMLR:v18:16-365} is utilized for the implementations of these benchmark undersampling and oversampling methods, and our implementations of SCL are mainly based on PyTorch \cite{paszke2019pytorch}. 

\begin{table}
\footnotesize
\centering
\caption{Description of 15 imbalanced datasets.}\label{data}
\begin{tabular}{ccccccc}
\hline
Data sets&Abbreviation&Size&Features&Class&Class Distribution&Data Repository\\
\hline
Glass0 &gl0 &214 & 8 & 2 & 144/70 &KEEL\\
Ecoli2&eo2&336 & 7 & 2 & 284/52 &KEEL\\
Yeast3&yt3&1484 & 8 & 2 & 1321/163 &KEEL\\
Yeast6&yt6&1484 & 8 & 2 & 1449/35 &KEEL\\
Vowel0&vw0&988 & 13 & 2 & 90/898 & KEEL\\
Haberman&hb&306 & 3 & 2 & 225/81 & KEEL\\
Yeast24&yt24&514 & 8 & 2 & 463/51 & KEEL\\
Pageblock0&pa0&5472& 10 & 2 & 4913/559 & KEEL\\
Scale Balance&bal&625 & 4 & 3 & 49/288/288&KEEL\\
Wine&wine&178 & 13 & 3 & 59/71/48 & KEEL\\
lymphography&lym&148 & 18 & 4 & 2/81/61/4 & KEEL\\
Glass&gla&214 &9 & 6 & 70/76/17/13/9/29 & UCI\\
Pageblocks&page&548 & 10 & 5 & 492/33/3/8/12 & KEEL\\
Dermatology&dt&358 & 34 & 6 & 111/60/71/48/48/20 & KEEL\\
\multirow{2}{*}{Penbased}&\multirow{2}{*}{pb} &\multirow{2}{*}{1100} & \multirow{2}{*}{16} & \multirow{2}{*}{10} & 115/114/114/106/114/ & \multirow{2}{*}{KEEL}\\
& & & & &106/105/115/105/106\\
\hline
\end{tabular}
\end{table}

\subsubsection{Evaluation Metrics}
\label{sec:metric}

We evaluate the model's performance with four metrics: accuracy, F-score, G-mean, and AUC \cite{bradley1997use}. Accuracy is a commonly used metric that summarizes the performance of a classification model as the proportion of correct predictions in the total number of predictions, but it is sensitive to data distributions. Accordingly, we supplement three other metrics to evaluate classifiers in skewed data fields. Single-class metrics are calculated for each class and are less susceptible to class imbalance, so they are suitable for imbalanced data classification. For example, precision metric measures the correctly predicted positive class sample and is computed using Eq. (\ref{precision}), and recall quantifies the proportion of correctly identified of all actual positive samples defined by Eq. (\ref{recall}). In general, precision and recall share an inverse relationship. In order to seek a balance between them, F-measure is proposed, as shown in Eq. (\ref{F1}). G-mean metric evaluates the degree of inductive bias between the accuracy of positive and negative classes.

{\small
\begin{equation}
\text { Accuracy }=\frac{T P+T N}{T P+F N+F P+T N} \\
\end{equation}
\begin{equation}
\label{precision}
\text { Precision }=\frac{T P}{T P+F P} \\
\end{equation}
\begin{equation}
\label{recall}
\text { Recall }=\frac{T P}{T P+F N}
\end{equation}
\begin{equation}
\label{F1}
\text { F }-\text { score }=\frac{2 \times \text { Recall } \times \text { Precision }}{\text { Precision }+\text { Recall }} 
\end{equation}
\begin{equation}
\text { G }-\text { mean }=\sqrt{\frac{T P}{T P+F N}} \times \sqrt{\frac{T N}{T N+F P}}
\end{equation}
}

Besides, we use the overall metric AUC. For imbalanced binary datasets, the ROC curve is plotted with TP against the FP where TP is on the y-axis and FP is on the x-axis. AUC metric converts this curve to a value, measuring the entire two-dimensional area underneath the ROC curve. For multi-class imbalanced problems, the MAUC metric averages the AUC value of all pairs of classes. This study calculates metrics for each label and finds their unweighted mean. The equation is given as follows:

\begin{equation}
\label{MAUC}
\text { MAUC }=\frac{1}{c(c-1)} \sum_{j=1}^{c} \sum_{k>j}^{c}(\operatorname{AUC}(j \mid k)+\operatorname{AUC}(k \mid j))
\end{equation}
where $c$ is the number of classes and $\operatorname{AUC}(j \mid k)$ is the AUC with class $j$ as the positive class and class $k$ as the negative class. In general, $\operatorname{AUC}(j \mid k) \neq \operatorname{AUC}(k \mid j))$ in the multiclass case. In our experiments, accuracy, F-measure, G-mean, and AUC are used together as assessment metric, in which accuracy and AUC are from sklearn.metrics \cite{pedregosa2011scikit}, F-measure and G-mean are from imblearn.metrics\cite{JMLR:v18:16-365}. For each metric, the greater the value, the better the performance.

\subsubsection{Parameters setting}
The detailed parameters for each model we construct are shown in Table \ref{parameters}. For all the datasets, the number of epochs of training contrastive network is 5000, the number of epochs of training linear classifier is 25. Adam is adopted as an optimizer, and the learning rate is 0.001. The number of TPE iterations is fixed at 75.

\begin{table}[h!] 
\scriptsize
\setlength{\tabcolsep}{1pt}
\centering
\caption{Parameters for SCL-TPE for imbalanced tabular datasets.}\label{parameters}
\begin{tabular}{ccccc}
\hline
\specialrule{0em}{0.6pt}{0.6pt}
\multirow{2}{*}{Data sets}& $\#$ of neurons in each&$\#$ of neurons in each&\multirow{2}{*}{Batch size}&$\tau$ chosen\\
&layer of extractor&  layer of linear classifier& &by TPE\\
\hline
Glass0 & (9,96,48)  & (48,24,2) & 160 &0.514\\
Ecoli2& (7,96,48)  & (48,10,2) & 128 &0.489\\
Yeast3& (8,128,64)  & (64,32,2) & 240 &0.857\\
Yeast6& (8,96,48) & (48,24,2) & 320 &0.947\\
Vowel0& (13,104,52) & (52,26,2) & 160 & 0.010\\
Haberman& (3,96,48) & (48,24,2) & 128 &0.153 \\
Yeast24& (8,128,64) &(64,32,2)&128& 0.245 \\
Pageblock0& (10,128,64) & (64,32,2) & 160 &0.348 \\
Scale Balance& (4,128,64)   & (64,32,3) &128&0.995\\
Wine& (13,200,100)  & (100,50,3)& 150&0.055 \\
lymphography& (18,128,64) & (62,32,4) & 150 & 0.854\\
Glass& (9,128,64)  & (64,32,6) & 128 & 0.352\\
Pageblocks& (10,128,64) & (64,32,5) & 128 & 0.122\\
Dermatology& (34,128,64) & (64,32,6) & 128 & 0.116\\
Penbased& (16,128,64) & (64,32,10) & 160 & 0.059\\
\hline
\end{tabular}
\label{parameters}
\end{table}

\subsection{Experimental results and analysis}
\subsubsection{Discussion of TPE}
\begin{figure}[H]
\centering
\subfigure[Glass 0. $\tau$ = 0.05]{
\includegraphics[width=5.7cm]{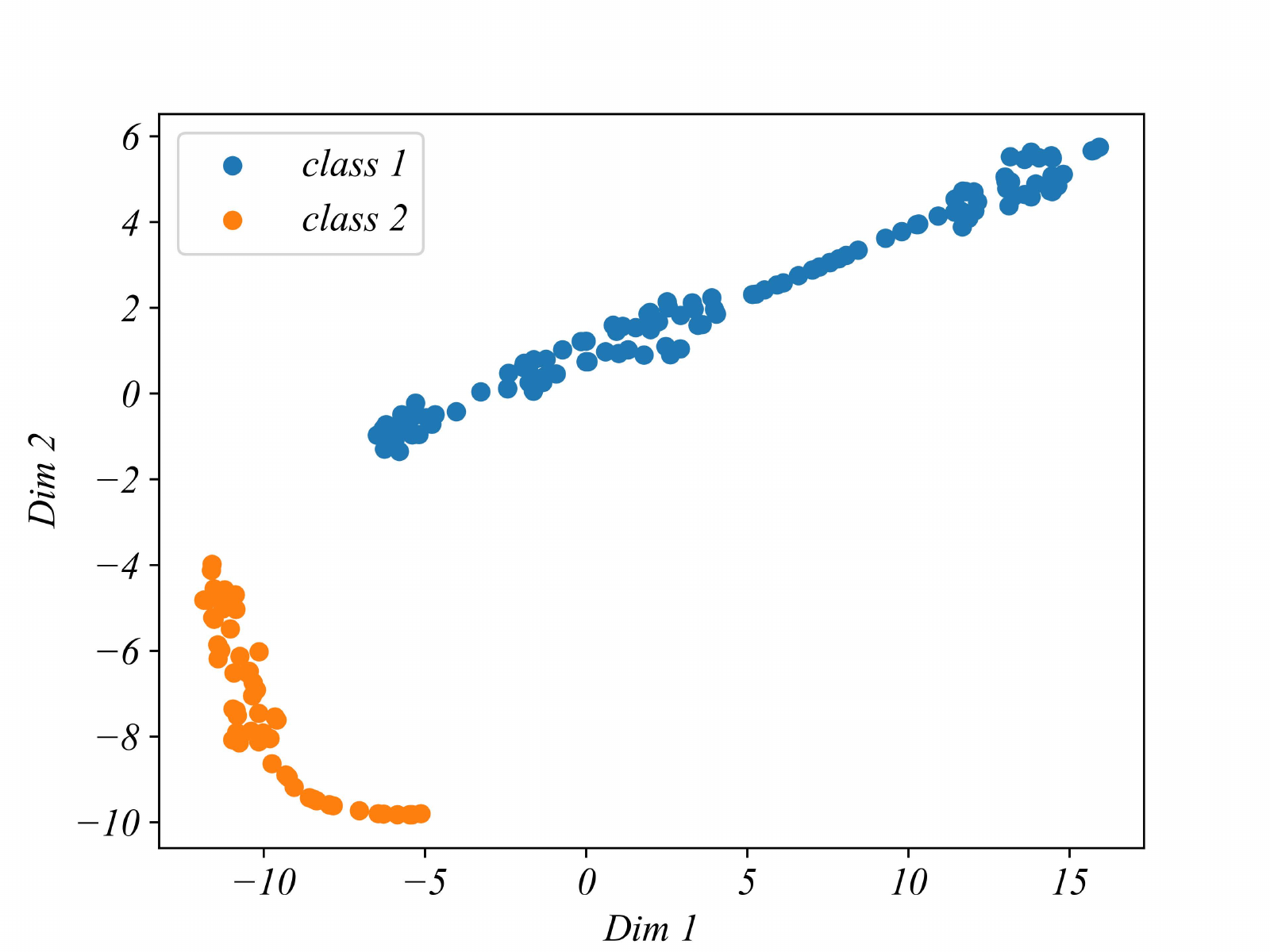}
}
\quad
\subfigure[Glass 0. $\tau$ = 0.6]{
\includegraphics[width=5.7cm]{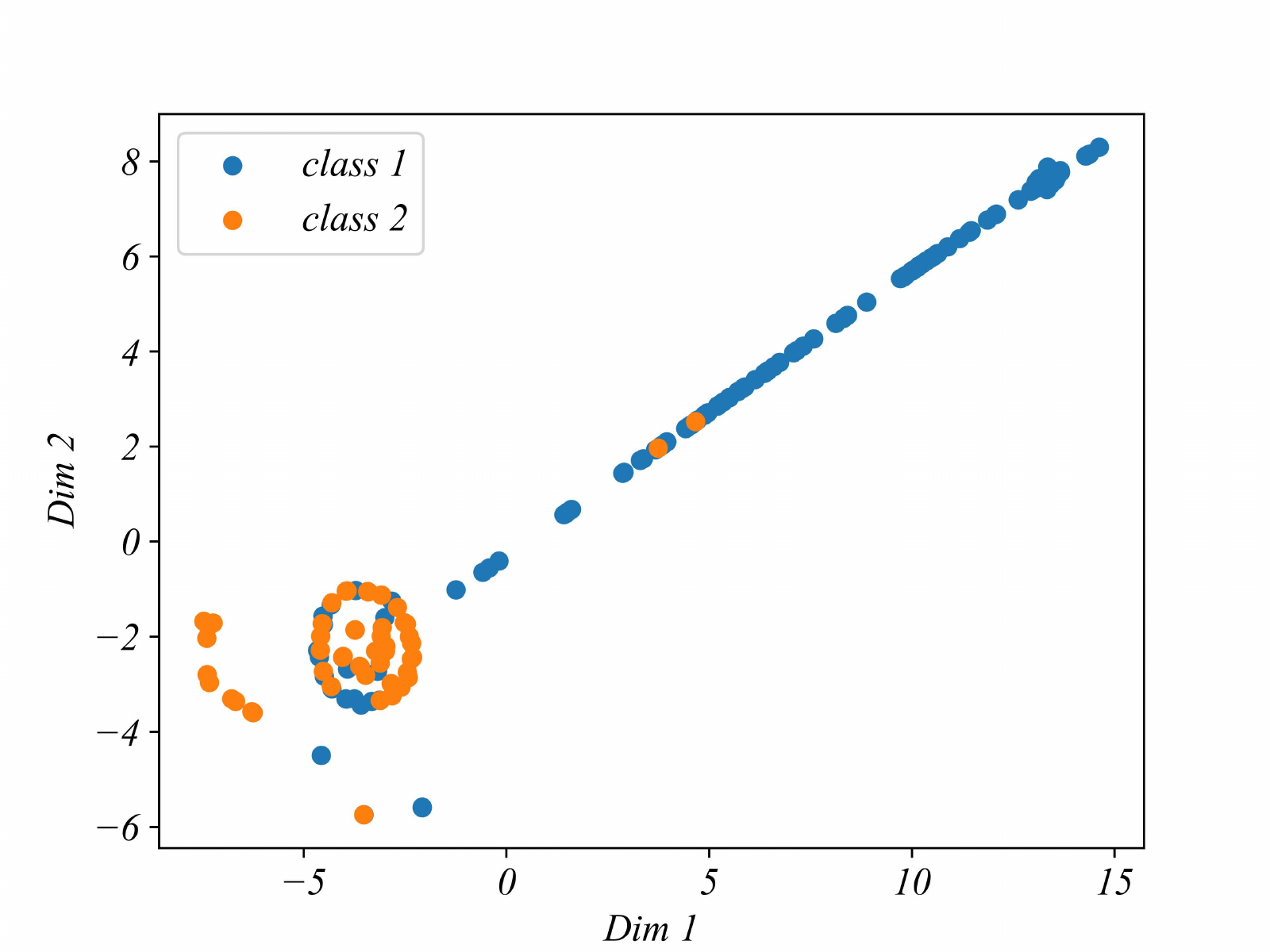}
}
\quad
\subfigure[Glass. $\tau$ = 0.05]{
\includegraphics[width=5.7cm]{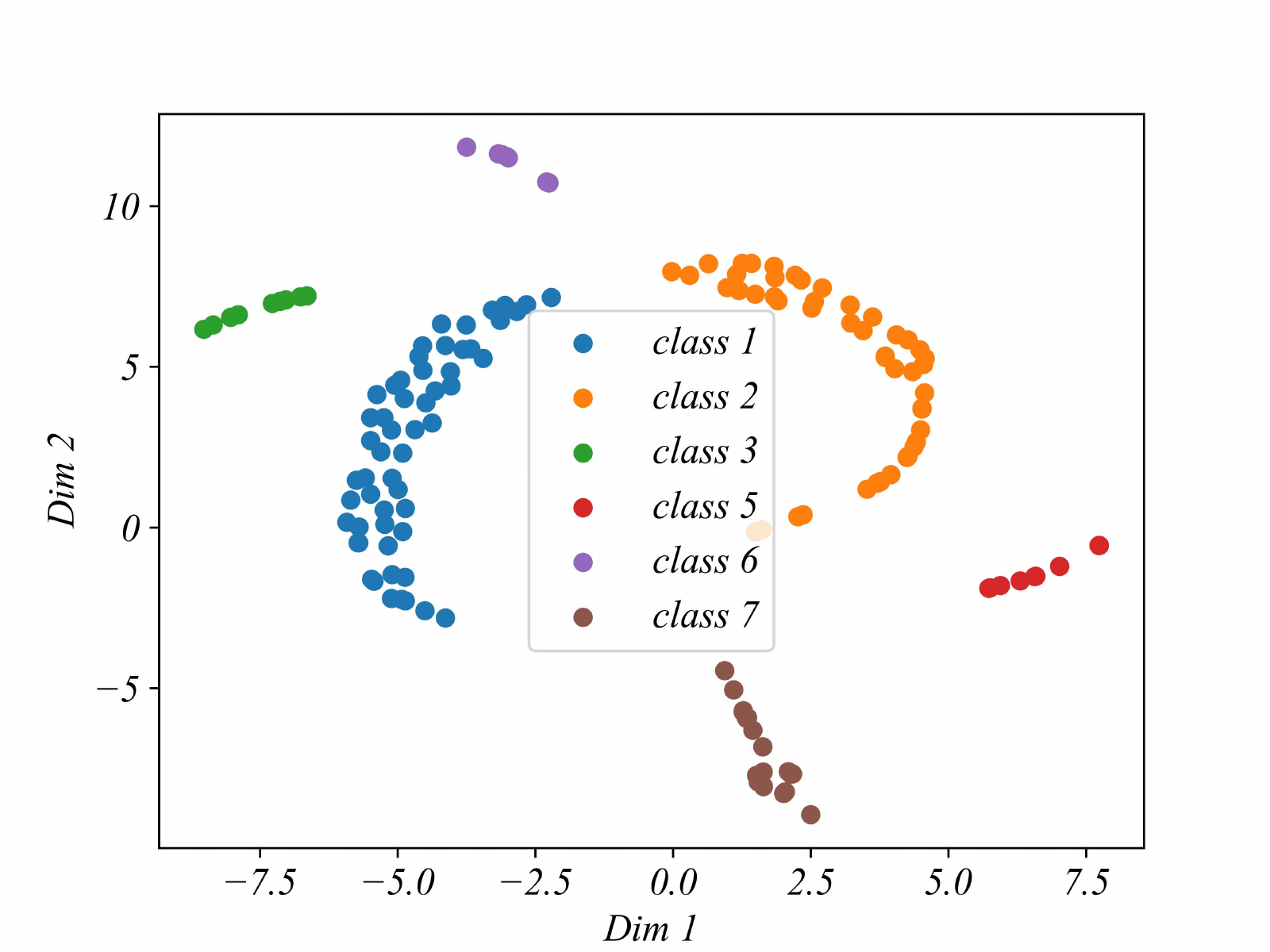}
}
\quad
\subfigure[Glass. $\tau$ = 0.34]{
\includegraphics[width=5.7cm]{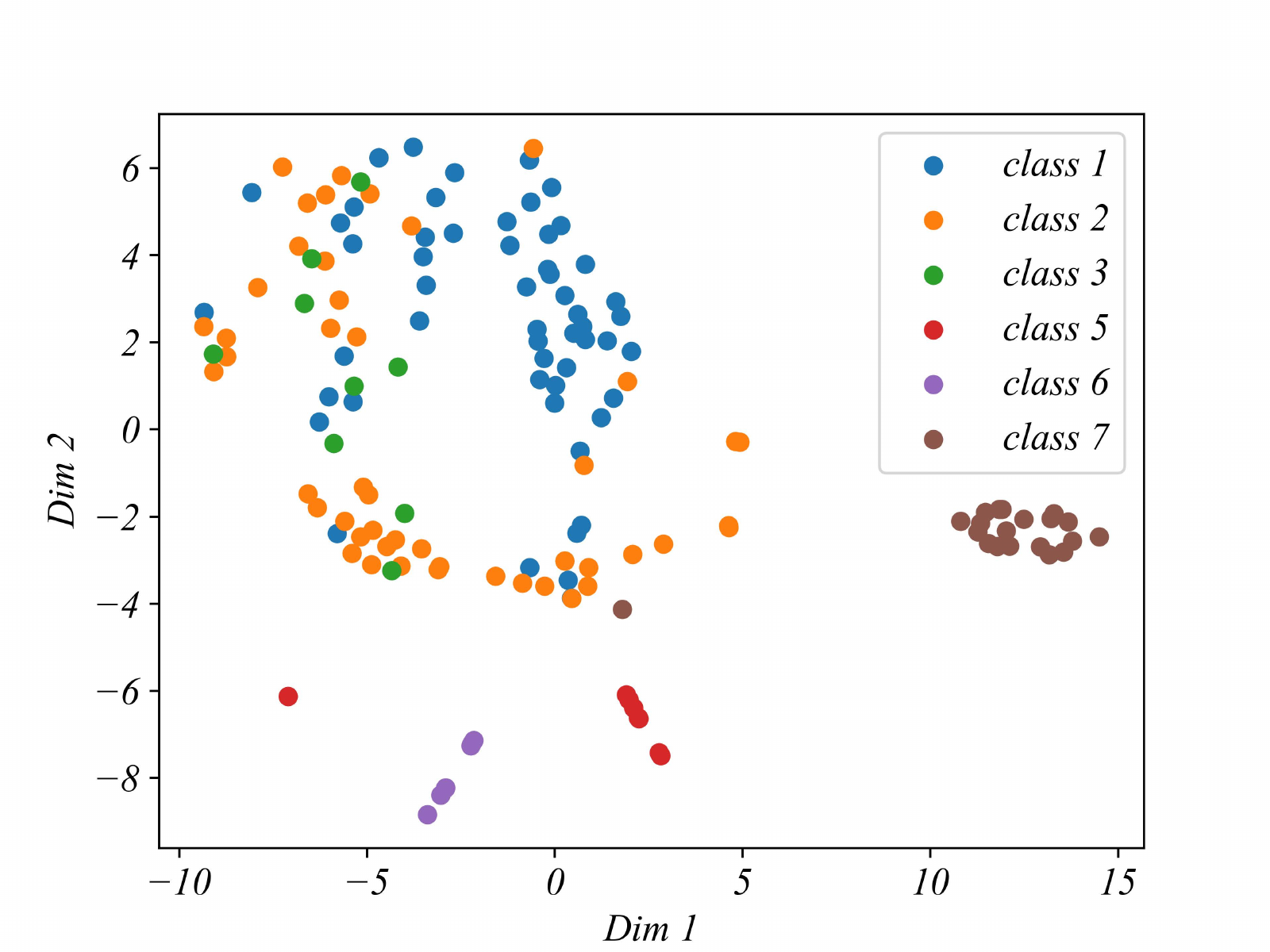}
}
\caption{The t-SNE embeddings of Glass0 and Glass.}\label{tsne-Glass}
\end{figure}

In this sub-experiment, we demonstrate that different values of hyper-parameter $\tau$ can lead to fluctuations in the model's performance. Bayesian optimization is also proved to be more effective and efficient in selecting a promising hyper-parameter than other HPO algorithms like GS, RS, and GA. We take one binary dataset and one multiclass dataset, Glass and Glass0, as examples. Fig. \ref{tsne-Glass} suggests that $\tau$ affects the quality of data embedding. For dataset Glass0, we observe that embeddings of $\tau$  = 0.05 present a more reasonable locally clustered and globally separated distribution, while the embeddings trained with $\tau$ = 0.6 are chaotic. This phenomenon can also be seen in the dataset Glass. We also evaluate the performances of classification results with different $\tau$ on Glass0, Glass. Tables \ref{Glass0 tau} and \ref{Glass tau} show the performance of classification results on the two datasets, respectively. We can see different $\tau$ values will lead to significant differences in final performance.

\begin{table}[h!] 
\scriptsize
\setlength{\tabcolsep}{2.45pt}
\centering
\caption{Performance comparison of models trained with different temperature on Glass0.}\label{Glass0 tau} 
\begin{tabular}{|l|c|c|c|c|c|c|c|c|c|c|c|c|}
\hline
\diagbox{Metrics}{$\tau$} &0.03&0.07&0.1&0.2&0.3&0.4&0.5&0.6&0.7&0.8&0.9&1.0\\
\hline
Accuracy & 0.721 &0.628&0.674&0.674&0.674& 0.651&0.767&0.697&0.744&0.628&0.628&0.674\\
F-measure &0.73 &0.61& 0.68 & 0.68 & 0.68 &0.66&0.77&0.59&0.75&0.64&0.64&0.54\\
G-mean &0.74  &0.48&    0.71&0.71&0.71&0.69&0.73&0.27&0.77&0.67&0.67&0.00\\
AUC & 0.738 &0.539&0.722&0.722&0.722& 0.704&0.735&0.536&0.773&0.687&0.687&0.500\\
\hline
\end{tabular}
\end{table}
 
 \begin{table}[h!] 
\scriptsize
\setlength{\tabcolsep}{2.45pt}
\centering
\caption{Performance comparison of models trained with different temperature on Glass.}\label{Glass tau} 
\begin{tabular}{|l|c|c|c|c|c|c|c|c|c|c|c|c|}
\hline
\diagbox{Metrics}{$\tau$} &0.03&0.07&0.1&0.2&0.3&0.4&0.5&0.6&0.7&0.8&0.9&1.0\\
\hline
Accuracy & 0.462 & 0.385 &0.400&0.569&0.215&0.323&0.185&0.538&0.262&0.262&0.308&0.400 \\
F-measure &0.38 &0.28 &0.28& 0.53&0.21&0.33&0.19&0.41&0.21&0.19&0.21&0.30\\
G-mean &0.39 & 0.39 & 0.33& 0.61 &0.33&0.48&0.29&0.37&0.30&0.33&0.27&0.40\\
AUC & 0.662& 0.697 & 0.725& 0.752&0.650&0.653&0.593&0.665&0.585&0.585&0.728&0.707\\
\hline
\end{tabular}
\end{table}

\begin{figure}[H]
\centering
\subfigure[Glass 0.]{
\includegraphics[width=5.7cm]{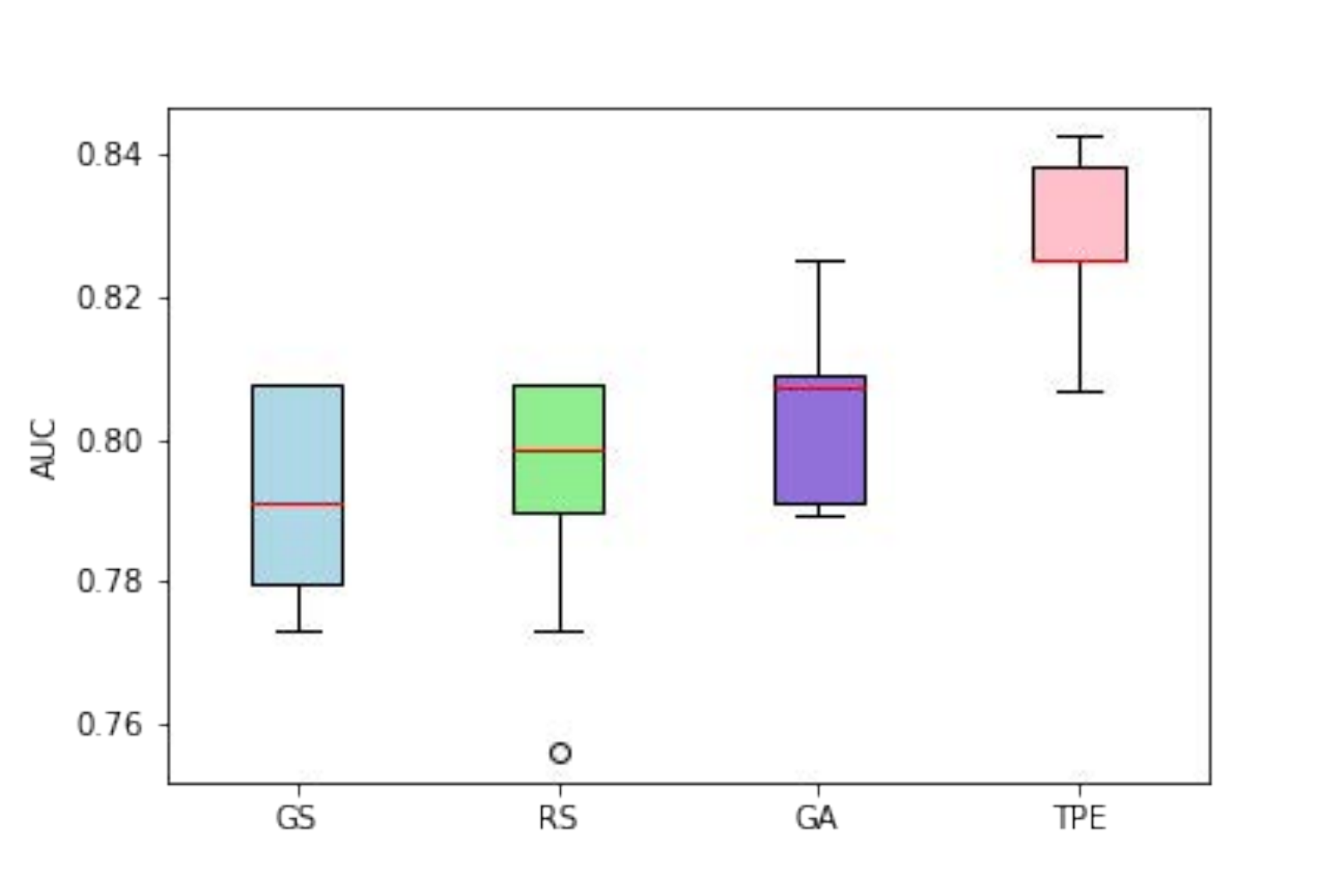}
}
\quad
\subfigure[Glass.]{
\includegraphics[width=5.7cm]{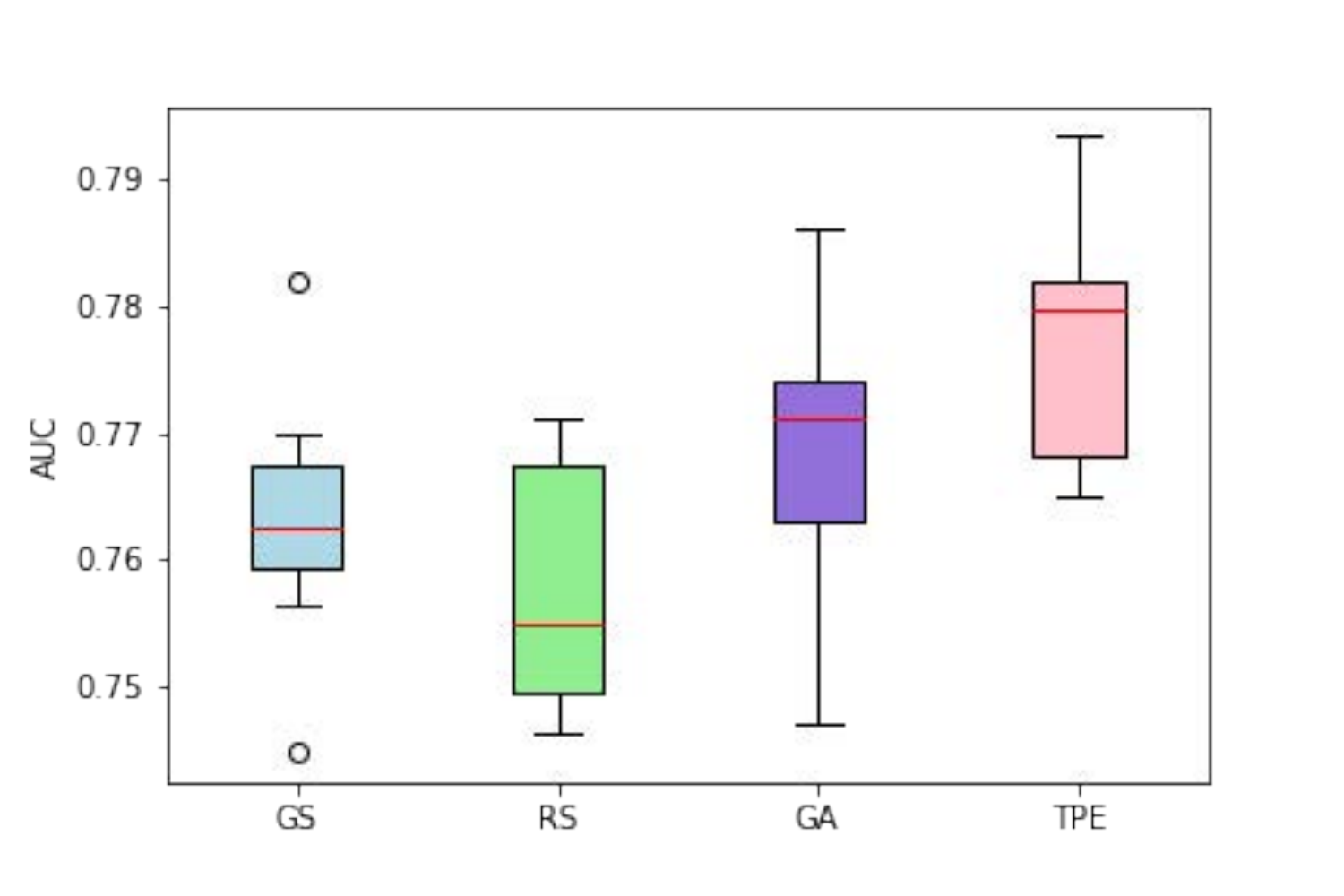}
}
\caption{Boxplot of accuracy over four HPO approach.}\label{Boxplot}
\end{figure}

This paper adopts the TPE, and the other three HPO strategies, GS, RS, and GA, are introduced as the baseline. The primary purpose of this sub-experiment is to evaluate the performance of the TPE. The RS and TPE optimizers are iterative processes with the number of iterations set to 75 so as to achieve a good balance between performance and complexity. In the GS, we changed the value of $\tau$ from 0 to 1 in increments of 0.02. In the GA, we set the number of iterations as 5 and the population size as 15. The population size here determines the number of trial solutions in each iteration. Fig. \ref{Boxplot} shows the boxplot of the AUC of different optimizers over two datasets, where the X-axis denotes the adopted methods, and the Y-axis represents the AUC. The median results are shown as the red line in the figure. We can see that TPE's optimal and medium results greatly exceed those of RS, GS, and GA. The effectiveness of the TPE is thus proved.

To test the efficiency of the TPE, we compared the performances and running times of these two methods with 75 iterations each. The population size of GA is set to 10. For Glass0, the results obtained by TPE and GA are the same. The running time of TPE is 4695s, and of GA is 23211s. For Glass, the accuracy of GA is 61.5$\%$ which is 3.1$\%$  lower than TPE; the F1 score is 54$\%$ which is 7.0$\%$ lower than TPE; the G-mean is 60$\%$ which is 9.0$\%$ lower than TPE; the AUC is 78.9$\%$ which is 0.2$\%$ higher than TPE. The running time of TPE is 5369s, and of GA is 26776s. We can see that GA results are slightly worse than TPE and take more time. As we mentioned in Section \ref{sec:intro}, that’s because, in each iteration, GA needs to train the model on multiple trial solutions. 

\subsubsection{SCL-TPE vs. state-of-the-art methods}
\begin{table}[!h]
\tiny
\centering
\tabcolsep=0.05cm
\caption{\centering Accuracy and F-measure for binary imbalanced data.}\label{binary1}
\begin{tabular}{cccccccccccccccccccccccc}
\toprule
\multirow{2}{*}{Data}& &\multicolumn{10}{c}{Accuracy}&\makebox[0.01\textwidth][c]{}&\multicolumn{10}{c}{F-measure}\\
\cline{3-12}
\cline{14-23}
\rule{0pt}{11pt}
&  & None&RUS&ENN&CNN&OSS&ROS&SMO&ADA&BSM&GAN&\makebox[0.01\textwidth][c]{}&\makebox[0.037\textwidth][c]{None}&\makebox[0.037\textwidth][c]{RUS}&\makebox[0.037\textwidth][c]{ENN}&\makebox[0.037\textwidth][c]{CNN}&\makebox[0.037\textwidth][c]{OSS}&\makebox[0.037\textwidth][c]{ROS}&\makebox[0.037\textwidth][c]{SMO}&\makebox[0.037\textwidth][c]{ADA}&\makebox[0.037\textwidth][c]{BSM}&\makebox[0.037\textwidth][c]{GAN}\\

\midrule
\multirow{6}{*}{gl0} & MLP &0.744&0.674&0.674&0.698&0.674&0.767&0.767&0.721&0.813&0.767&& 0.75&0.54&0.54&0.70&0.54&0.76&0.76&0.72&0.82&0.76\\
&SVM &0.674&0.581&0.605&0.326&0.674&0.581&0.581&0.581&0.581&0.628&& 0.54&0.57&0.60&0.16&0.54&0.57&0.57&0.57&0.57&0.63 \\
&KNN &0.767&0.744&0.698&0.791&0.767&0.744&0.767&0.721&0.767&0.767&& 0.77&0.75&0.71&0.80&0.77&0.75&0.77&0.73&0.77&0.77\\
&DTC &0.605&0.651&0.651&0.558&0.628&0.651&0.674&0.628&0.581&0.674&& 0.61&0.66&0.66&0.57&0.64&0.66&0.68&0.64&0.54&0.68\\
&RFC&0.674&0.674&0.698&0.721& 0.628&0.628&0.651&0.628&0.674&0.651&& 0.66&0.68&0.68&0.72&0.61&0.58&0.63&0.55&0.63&0.66\\
&SCL-TPE&\textbf{0.837} & & & & & & & & & &&\textbf{0.84}\\
\specialrule{0em}{0.6pt}{0.6pt}
\hline
\specialrule{0em}{0.6pt}{0.6pt}
\multirow{6}{*}{eo2} & MLP &0.744&0.674&0.674&0.698&0.674&0.767&0.767&0.721&0.813&0.767&& 0.75&0.54&0.54&0.70&0.54&0.76&0.76&0.72&0.82&0.76\\
&SVM &0.674&0.581&0.605&0.326&0.674&0.581&0.581&0.581&0.581&0.628&& 0.54&0.57&0.60&0.16&0.54&0.57&0.57&0.57&0.57&0.63 \\
&KNN &0.767&0.744&0.698&0.791&0.767&0.744&0.767&0.721&0.767&0.767&& 0.77&0.75&0.71&0.80&0.77&0.75&0.77&0.73&0.77&0.77\\
&DTC &0.605&0.651&0.651&0.558&0.628&0.651&0.674&0.628&0.581&0.674&& 0.61&0.66&0.66&0.57&0.64&0.66&0.68&0.64&0.54&0.68\\
&RFC&0.674&0.674&0.698&0.721&0.628&0.628&0.651&0.628&0.674&0.651&&0.66&0.68&0.68&0.72&0.61&0.58&0.63&0.55&0.63&0.66\\
&SCL-TPE&\textbf{0.837} & & & & & & & & & &&\textbf{0.84}\\
\specialrule{0em}{0.6pt}{0.6pt}
\hline
\specialrule{0em}{0.6pt}{0.6pt}
\multirow{6}{*}{yt3} & MLP &0.912&0.823&0.838&0.838&0.912&0.897&0.838&0.926&0.941&0.912&&0.91&0.84&0.76&0.76&0.91&0.90&0.85&0.93&0.94&0.91\\
&SVM &0.867&0.794&0.912&0.691&0.882&0.779&0.765&0.750&0.691&0.735&&0.83&0.82&0.91&0.73&0.85&0.81&0.79&0.78&0.73&0.77\\
&KNN &0.911&0.809&0.868&0.735&0.897&0.868&0.882&0.853&0.897&0.882&& 0.91&0.83&0.88&0.77&0.90&0.88&0.89&0.86&0.90&0.89\\
&DTC &0.853&0.721&0.912&0.824&0.779&0.853&0.794&0.868&0.912&0.912&& 0.86&0.76&0.91&0.84&0.81&0.85&0.81&0.87&0.91&0.92\\
&RFC &0.941&0.794&0.926&0.882&0.926&0.912&0.926&0.912&0.926&0.926&& 0.94&0.82&0.93&0.89&0.93&0.91&0.93&0.91&0.93&0.93\\
&SCL-TPE&\textbf{0.956} & & & & & & & & & &&\textbf{0.96}\\
\specialrule{0em}{0.6pt}{0.6pt}
\hline
\specialrule{0em}{0.6pt}{0.6pt}

\multirow{6}{*}{yt6} & MLP &0.976&0.680&0.976&0.976&0.976&0.976&0.986&0.979&\textbf{0.989}&0.973&& 0.96&0.79&0.96&0.96&0.96&0.96&0.98&0.98&\textbf{0.99}&0.97\\
&SVM &0.976&0.811&0.976&0.976&0.976&0.882&0.892&0.886&0.902&0.943&&0.96&0.88&0.96&0.96&0.96&0.92&0.93&0.92&0.93&0.96\\
&KNN &0.983&0.862&0.980&0.983&0.983&0.963&0.953&0.956&0.970&0.973&&0.98&0.91&0.98&0.98&0.98&0.97&0.96&0.97&0.97&0.97\\
&DTC &0.966&0.687&0.963&0.939&0.963&0.969&0.872&0.862&0.956&0.970&&0.97&0.79&0.97&0.95&0.97&0.96&0.91&0.90&0.95&0.97\\
&RFC&0.979&0.919&0.987&0.976&0.976&0.976&0.976&0.976&0.976&0.983&& 0.97&0.94&0.98&0.96&0.96&0.96&0.96&0.96&0.96&0.98\\
&SCL-TPE&0.976& & & & & & & & & &&0.98\\
\specialrule{0em}{0.6pt}{0.6pt}
\hline
\specialrule{0em}{0.6pt}{0.6pt}

\multirow{6}{*}{vw0} & MLP &0.909&0.949&0.955&0.965&0.955&0.944&0.995&\textbf{1.0}&\textbf{1.0}&\textbf{1.0}&&0.87&0.95&0.95&0.96&0.95&0.94&\textbf{1.0}&\textbf{1.0}&\textbf{1.0}&\textbf{1.0}\\
&SVM &0.909&0.884&0.909&0.091&0.909&0.914&0.914&0.924&0.914&0.960&& 0.87&0.90&0.87&0.02&0.87&0.93&0.93&0.93&0.93&0.96\\
&KNN &\textbf{1.0}&0.889&0.985&0.813&0.995&0.997&0.995&0.989&0.997&0.995&&\textbf{1.0}&0.91&0.99&0.85&\textbf{1.0}&\textbf{1.0} &\textbf{1.0}&0.99&\textbf{1.0}&\textbf{1.0}\\
&DTC &0.939&0.924&0.970&0.894&0.944&0.985&0.975&0.975&0.965&0.985&&0.95&0.93&0.97&0.91&0.95&0.98&0.98&0.97&0.97&0.99\\
&RFC&0.975&0.954&0.995&0.914&\textbf{1.0}&\textbf{1.0}&0.995&0.995&0.989&\textbf{1.0}&&0.98&0.96&0.99&0.93&\textbf{1.0}&\textbf{1.0}&\textbf{1.0}&\textbf{1.0}&0.99&\textbf{1.0}\\
&SCL-TPE&\textbf{1.0} & & & & & & & & & &&\textbf{1.0}\\
\specialrule{0em}{0.6pt}{0.6pt}
\hline
\specialrule{0em}{0.6pt}{0.6pt}

\multirow{6}{*}{hb} & MLP &0.677&0.516&0.725&\textbf{0.742}&0.613 &0.725&0.629&0.677&0.581&0.677 &&0.61&0.54&0.61&0.65&0.63&0.61&0.65&0.69&0.60&0.67\\
&SVM &0.725&0.677&0.645&0.677&0.661&0.581&0.613&0.597&0.565&0.613&&0.61&0.67&0.63&0.64&0.62&0.59&0.61&0.60&0.57&0.62\\
&KNN &0.677&0.548&0.597&0.677&0.645&0.645&0.613&0.613&0.613&0.694&&0.67&0.57&0.62&0.68&0.65&0.66&0.63&0.63&0.63&\textbf{0.71}\\
&DTC &0.581&0.548&0.484&0.597&0.677&0.677&0.661&0.629&0.581&0.677&& 0.59&0.57&0.50&0.61&0.68&0.68&0.67&0.63&0.60&0.66\\
&RFC&0.677&0.613&0.597&0.694&0.677&0.661&0.597&0.613&0.597&0.613&&0.67&0.63&0.62&0.70&0.69&0.66&0.61&0.63&0.61&0.62\\
&SCL-TPE&0.645 & & & & & & & & & &&0.66\\
\specialrule{0em}{0.6pt}{0.6pt}
\hline
\specialrule{0em}{0.6pt}{0.6pt}

\multirow{6}{*}{yt24} & MLP &0.913&0.932&0.922&0.951&0.961&0.942&0.903&0.903&0.903&0.922&& 0.91&0.94&0.92&0.95&0.96&0.94&0.86&0.86&0.86&0.93\\
&SVM &0.903&0.864&0.922&0.854&0.903&0.883&0.893&0.883&0.883&0.951&&0.86&0.88&0.90&0.88&0.86&0.90&0.90&0.90&0.90&0.95\\
&KNN &0.961&0.922&0.961&0.903&0.961&0.942&0.932&0.942&0.942&0.961&&0.96&0.93&0.96&0.91&0.96&0.94&0.94&0.95&0.94&0.96\\
&DTC &0.903 &0.893&0.903&0.951&0.932&0.942&0.932&0.893&0.951&0.922&&0.91&0.90&0.91&0.95&0.93&0.94&0.93&0.90&0.95&0.93\\
&RFC&0.951&0.903&0.932&0.942&0.951&0.951&0.961&0.961&0.961&0.951&&0.95&0.91&0.93&0.94&0.95&0.95&0.96&0.96&0.96&0.95\\
&SCL-TPE&\textbf{0.971} & & & & & & & & & &&\textbf{0.97}\\
\specialrule{0em}{0.6pt}{0.6pt}
\hline
\specialrule{0em}{0.6pt}{0.6pt}

\multirow{6}{*}{pa0} & MLP &0.923&0.815&0.906&0.687&0.898&0.910&0.866&0.898&0.915&0.841&& 0.93&0.85&0.92&0.75&0.85&0.92&0.89&0.85&0.92&0.87\\
&SVM &0.919&0.932&0.929&0.681&0.920&0.928&0.927&0.881&0.881&0.909&& 0.90&0.93&0.91&0.74&0.90&0.93&0.93&0.90&0.90&0.92\\
&KNN &0.947&0.920&0.937&0.940&0.947&0.921&0.919&0.908&0.920&0.958&&0.94&0.93&0.94&0.94&0.94&0.93&0.92&0.92&0.93&0.96\\
&DTC &0.816&0.289&0.769&0.728&0.829&0.898&0.817&0.628&0.533&0.817&& 0.85&0.34&0.81&0.78&0.86&0.91&0.85&0.64&0.62&0.85\\
&RFC &0.674&0.772&0.683&0.876&0.873&\textbf{0.961}&0.788&0.601&0.595&0.905&&0.66&0.82&0.75&0.89&0.89&\textbf{0.96}&0.83&0.68&0.67&0.92\\
&SCL-TPE&0.918 & & & & & & & & & &&0.93\\

\bottomrule
\end{tabular}
\end{table}

\begin{table}[!h]
\tiny
\centering
\tabcolsep=0.05cm
\caption{\centering G-mean and AUC for binary imbalanced data.}\label{binary2}
\begin{tabular}{cccccccccccccccccccccccc}
\toprule
\multirow{2}{*}{Data}& &\multicolumn{10}{c}{G-mean}&\makebox[0.01\textwidth][c]{}&\multicolumn{10}{c}{AUC}\\
\cline{3-12}
\cline{14-23}
\rule{0pt}{11pt}
&  & None&RUS&ENN&CNN&OSS&ROS&SMO&ADA&BSM&GAN&\makebox[0.01\textwidth][c]{}&\makebox[0.037\textwidth][c]{None}&\makebox[0.037\textwidth][c]{RUS}&\makebox[0.037\textwidth][c]{ENN}&\makebox[0.037\textwidth][c]{CNN}&\makebox[0.037\textwidth][c]{OSS}&\makebox[0.037\textwidth][c]{ROS}&\makebox[0.037\textwidth][c]{SMO}&\makebox[0.037\textwidth][c]{ADA}&\makebox[0.037\textwidth][c]{BSM}&\makebox[0.037\textwidth][c]{GAN}\\

\midrule
\multirow{6}{*}{gl0} & MLP &0.71&0.00&0.00&0.74&0.00&0.67&0.70&0.67&0.81&0.70&&0.717&0.500&0.500&0.757&0.500&0.698&0.717&0.682&0.806&0.717\\
&SVM &0.00&0.62&0.64&0.00&0.00&0.62&0.62&0.62&0.62&0.67&&0.500&0.690&0.706&0.500&0.500&0.689&0.689&0.689&0.689&0.706\\
&KNN &0.75&0.74&0.73&0.82&0.75&0.74&0.75&0.72&0.77&0.75&& 0.754&0.736&0.739&0.826&0.754&0.744&0.754&0.719&0.772&0.754\\
&DTC &0.64&0.67&0.68&0.59&0.65&0.69&0.71&0.65&0.34&0.68&&0.651&0.667&0.685&0.599&0.650&0.704&0.722&0.650&0.468&0.685\\
&RFC&0.54&0.67&0.55&0.67&0.48&0.35&0.49&0.25&0.44&0.69&&0.592&0.666&0.610&0.682&0.539&0.502&0.557&0.484&0.555&0.704\\
&SCL-TPE&\textbf{0.84} & & & & & & & & & &&\textbf{0.842}\\
\specialrule{0em}{0.6pt}{0.6pt}
\hline
\specialrule{0em}{0.6pt}{0.6pt}
\multirow{6}{*}{eo2} & MLP &0.78&0.89&0.00&0.00&0.83&0.78&0.75&0.84&0.85&0.83&&0.801&0.895&0.500&0.500&0.837&0.792&0.757&0.846&0.855& 0.837\\
&SVM &0.43&0.84&0.78&0.79&0.52&0.83&0.79&0.78&0.74&0.80&& 0.591&0.841&0.801&0.816&0.636&0.832&0.786&0.778&0.742&0.805\\
&KNN &0.87&0.81&0.85&0.83&0.86&0.85&0.86&0.84&0.86&0.86&& 0.874&0.813&0.848&0.842&0.865&0.848&0.856&0.839&0.865&0.856\\
&DTC &0.75&0.76&0.87&0.86&0.83&0.71&0.77&0.76&0.87&0.91&& 0.766&0.760&0.874&0.858&0.832&0.729&0.767&0.774&0.874&0.911\\
&RFC &0.85&0.80&0.84& 0.89&0.84&0.78&0.84&0.78&0.84&0.88&&0.855&0.804&0.846&0.893&0.846&0.801&0.846&0.801&0.846&0.884&\\
&SCL-TPE&\textbf{0.90} & & & & & & & & & &&\textbf{0.900}\\
\specialrule{0em}{0.6pt}{0.6pt}
\hline
\specialrule{0em}{0.6pt}{0.6pt}
\multirow{6}{*}{yt3} & MLP &0.00&0.86&0.00&0.86&0.92&0.81&0.78&0.00&0.92&0.81&&0.500&0.871&0.500&0.864&0.924&0.818&0.784&0.500&0.916& 0.809\\
&SVM &0.00&0.87&0.39&0.72&0.17&0.89&0.89&0.83&0.80&0.72&&0.500&0.879&0.572&0.746&0.513&0.892&0.892&0.841&0.816&0.744\\
&KNN &0.83&0.84&0.86&0.84&0.83&0.89&0.91&0.89&0.90&0.84&&0.835&0.845&0.860&0.845&0.835&0.892&0.907&0.898&0.905&0.848\\
&DTC &0.89&0.83&0.82&0.87&0.80&0.84&0.81&0.76&0.75&0.90&&0.839&0.833&0.826&0.867&0.807&0.845&0.814&0.758&0.752&0.898\\
&RFC &0.84&0.90&0.87&0.93&0.82&0.85&0.91&0.89&0.88&0.90&&0.847&0.909&0.871&0.936&0.828&0.852&0.907&0.890&0.881&0.902\\
&SCL-TPE&\textbf{0.95} & & & & & & & & & &&\textbf{0.955}\\
\specialrule{0em}{0.6pt}{0.6pt}
\hline
\specialrule{0em}{0.6pt}{0.6pt}

\multirow{6}{*}{yt6} & MLP &0.00&0.76&0.00&0.00&0.00&0.00&0.65&0.65&0.84&0.75&& 0.500&0.767&0.500&0.500&0.500&0.500&0.714&0.711&0.855&0.777 \\
&SVM &0.00&0.90&0.00&0.00&0.00&0.87&0.87&0.87&0.88&0.90&& 0.500&0.903&0.500&0.500&0.500&0.869&0.875& 0.872&0.880&0.910\\
&KNN &0.65&0.86&0.75&0.84&0.75&0.91&0.90&0.91&0.91&0.75&&0.713&0.860&0.781&0.852&0.782&0.911&0.906&0.908&0.915&0.777\\
&DTC &0.75&0.76&0.83&0.37&0.65&0.00&0.00&0.35&0.00&0.75&&0.774&0.770&0.842&0.551&0.702&0.497&0.446&0.511&0.490&0.775\\
&RFC &0.38&0.89&0.65&0.00&0.00&0.00&0.00&0.00&0.00&0.53&&0.571&0.889&0.714&0.500&0.500&0.500&0.500&0.500&0.500&0.643\\
&SCL-TPE&\textbf{0.92} & & & & & & & & &  &&\textbf{0.918}\\
\specialrule{0em}{0.6pt}{0.6pt}
\hline
\specialrule{0em}{0.6pt}{0.6pt}

\multirow{6}{*}{vw0} & MLP &0.00&0.78&0.78&0.78&0.78&0.77&\textbf{1.0}&\textbf{1.0}&\textbf{1.0}&\textbf{1.0}&& 0.500&0.797&0.806&0.806&0.800&0.794&0.997&\textbf{1.0}&\textbf{1.0}&\textbf{1.0}\\
&SVM &0.00&0.91&0.00&0.00&0.00&0.95&0.95&0.96&0.95&0.84&&0.500&0.911&0.500&0.500&0.500&0.953&0.953&0.958&0.953&0.853\\
&KNN &\textbf{1.0}&0.94&0.99&0.89&\textbf{1.0}&\textbf{1.0}&\textbf{1.0}&0.99&\textbf{1.0}&\textbf{1.0}&&\textbf{1.0}&0.992&0.897&0.997&0.997&0.997&0.997&0.994 &0.997&0.997\\
&DTC &0.97&0.96&0.93&0.92&0.94&0.91&0.96&0.91&0.90&0.97&&0.967&0.958&0.933&0.917&0.944&0.917&0.961&0.911&0.906&0.967\\
&RFC&0.99&0.97&0.97&0.95&\textbf{1.0}&\textbf{1.0}&\textbf{1.0}&\textbf{1.0}&0.99&\textbf{1.0}&& 0.986&0.975&0.972&0.953&\textbf{1.0}&\textbf{1.0}&0.997&0.997&0.994&\textbf{1.0} \\
&SCL-TPE&\textbf{1.0} & & & & & & & & & &&\textbf{1.0}\\
\specialrule{0em}{0.6pt}{0.6pt}
\hline
\specialrule{0em}{0.6pt}{0.6pt}

\multirow{6}{*}{hb} & MLP &0.23&0.42&0.00&0.24&0.56&0.00&0.59&0.65&0.56&0.53 &&0.485&0.447&0.500&0.529&0.568&0.500&0.598&0.659&0.565&0.576\\
&SVM &0.00&0.53&0.43&0.39&0.32&0.45&0.46&0.46&0.44&0.50&& 0.500&0.576&0.518&0.522&0.492&0.492&0.514&0.503&0.480&0.532\\
&KNN &0.53&0.58&0.61&0.60&0.52&0.63&0.64&0.60&0.62&\textbf{0.68}&&0.576&0.579&0.612&0.613&0.554&0.627&0.642&0.605&0.624&0.679\\
&DTC &0.45&0.58&0.53&0.55&0.57&0.60&0.61&0.51&0.52&0.49&&0.492&0.579&0.553&0.558&0.595&0.613&0.620&0.543&0.528&0.558\\
&RFC&0.53&0.60&0.61&0.60&0.62&0.52&0.49&0.56&0.49&0.53&&0.576&0.605&0.612&0.624&0.631&0.565&0.521&0.569&0.521&0.550\\
&SCL-TPE&\textbf{0.68} & & & & & & & & & &&\textbf{0.682}\\
\specialrule{0em}{0.6pt}{0.6pt}
\hline
\specialrule{0em}{0.6pt}{0.6pt}

\multirow{6}{*}{yt24} & MLP &0.75&0.92&0.76&0.88&0.88&0.77&0.00&0.00&0.00&0.87&& 0.773&0.918&0.778&0.884&0.889&0.789&0.500&0.500&0.500&0.868\\
&SVM &0.00&0.83&0.45&0.83&0.00&0.90&0.85&0.89&0.84&0.83&&0.500&0.835&0.600&0.830&0.500&0.846&0.852&0.891&0.846&0.839\\
&KNN &0.83&0.87&0.83&0.80&0.83&0.87&0.87&0.92&0.87&0.83&&0.845&0.868&0.845&0.812&0.845&0.878&0.873&0.923&0.878&0.845\\
&DTC&0.80&0.68&0.75&0.83&0.82&0.77&0.70&0.74&0.77&0.81&&0.812&0.718&0.768&0.839&0.828&0.789&0.739&0.762&0.795&0.823\\
&RFC&0.83&0.86&0.82&0.87&0.83&0.83&0.88&0.88&0.88&0.83&&0.839&0.857&0.828&0.878&0.839&0.839&0.889&0.889&0.889&0.839\\
&SCL-TPE&\textbf{0.98} & & & & & & & & & &&\textbf{0.984}\\
\specialrule{0em}{0.6pt}{0.6pt}
\hline
\specialrule{0em}{0.6pt}{0.6pt}

\multirow{6}{*}{pa0} & MLP &0.93&0.89&0.93&0.79&0.00&0.93&0.91&0.00&0.93&0.89 &&0.926&0.893&0.936&0.810&0.500&0.930&0.913&0.500&0.933&0.892\\
&SVM &0.48&0.86&0.57&0.77&0.48&0.88&0.88&0.87&0.87&0.85&& 0.615&0.863&0.660&0.775&0.615&0.885&0.884&0.867&0.871&0.854\\
&KNN &0.77&0.86&0.57&0.77&0.48&0.86&0.86&0.87&0.87&0.86&&0.796&0.863&0.660&0.775&0.615&0.861&0.860&0.870&0.872&0.866\\
&DTC &0.80&0.45&0.86&0.63&0.83&0.89&0.85&0.65&0.67&0.84&&0.799&0.592&0.863&0.639&0.830&0.888&0.847&0.650&0.704&0.839\\
&RFC &0.54&0.86&0.80&0.92&0.90&\textbf{0.94}&0.85&0.75&0.74&0.92&& 0.592&0.869&0.812&0.927&0.902&0.939&0.854&0.778&0.766&0.915\\
&SCL-TPE&\textbf{0.94} & & & & & & & & & &&\textbf{0.942}\\
\bottomrule
\end{tabular}
\end{table}

The following experiment is organized in the way below. We consider two cases: binary-class imbalanced data classification and multi-class imbalanced data classification. For each case, we compare SCL-TPE with other competitive methods to validate the superiority of the proposed method. The data are divided into two parts during the experiment, including the training and test datasets. We calculate and report the results based on the test part.
In imbalanced binary classification, the experimental results in Tables \ref{binary1} and \ref{binary2} present an overwhelming improvement of the proposed method over its competitors. In particular, our approach provides the best performance on all eight datasets when considering G-mean and AUC as performance measures. If we take the dataset Glass0 as an example, the proposed method yields an accuracy result of 0.837, which is 2.4$\%$ better than the second best method (BSMOTE with MLP), F-measure of 0.84 which is 2$\%$ higher than the second best method, G-mean of 0.84 which is 3$\%$ higher than the second best method, and AUC 0.842 which is 3.6$\%$ superior to the second best method. Regarding accuracy and F1 score, SCL-TPE achieves the optimal value on the five datasets. We found the results are not as good as other methods on the remaining three datasets because other methods tend to classify minority samples into the majority class. For example, in the Haberman dataset, OSS with MLP classifies all samples as the majority. However, in reality, minority groups are usually more important and need to be accurately identified.

\begin{table}[!h]
\tiny
\centering
\tabcolsep=0.05cm
\caption{\centering Accuracy and F-measure for multi-class imbalanced data.}\label{multi1}
\begin{tabular}{cccccccccccccccccccccccc}
\toprule
\multirow{2}{*}{Data}& &\multicolumn{10}{c}{Accuracy}&\makebox[0.01\textwidth][c]{}&\multicolumn{10}{c}{F-measure}\\
\cline{3-12}
\cline{14-23}
\rule{0pt}{11pt}
&  & None&RUS&ENN&CNN&OSS&ROS&SMO&ADA&BSM&GAN&\makebox[0.01\textwidth][c]{}&\makebox[0.037\textwidth][c]{None}&\makebox[0.037\textwidth][c]{RUS}&\makebox[0.037\textwidth][c]{ENN}&\makebox[0.037\textwidth][c]{CNN}&\makebox[0.037\textwidth][c]{OSS}&\makebox[0.037\textwidth][c]{ROS}&\makebox[0.037\textwidth][c]{SMO}&\makebox[0.037\textwidth][c]{ADA}&\makebox[0.037\textwidth][c]{BSM}&\makebox[0.037\textwidth][c]{GAN}\\

\midrule
\multirow{6}{*}{bal} & MLP &0.944&0.928&0.888&0.976&0.944&0.984&0.976&0.896&0.992&0.960&&0.93&0.94&0.88&0.97&0.93&0.98&0.97&0.90&0.99&0.95\\
&SVM &0.888&0.728&0.896&0.920&0.880&0.880&0.888&0.896&0.888&0.888&&0.90&0.79&0.86&0.88&0.85&0.90&0.91&0.91&0.91&0.85\\
&KNN &0.776&0.752&0.792&0.768&0.784&0.696&0.712&0.728&0.736&0.784&& 0.78&0.80&0.81& 0.82&0.78&0.73&0.74&0.76&0.76&0.78\\
&DTC &0.776&0.656&0.776&0.784&0.768&0.784&0.768&0.760& 0.784&0.776&& 0.80&0.73&0.80&0.75&0.80&0.75&0.79&0.77&0.80&0.80\\
&RFC&0.789&0.760&0.776&0.792&0.792&0.768&0.792&0.784&0.776&0.792&&0.78&0.78&0.78&0.80&0.77&0.74&0.79&0.78&0.77&0.79\\
&SCL-TPE&\textbf{1.0} & & & & & & & & & &&\textbf{1.0}\\
\specialrule{0em}{0.6pt}{0.6pt}
\hline
\specialrule{0em}{0.6pt}{0.6pt}
\multirow{6}{*}{wine} & MLP &0.567&0.333&0.694&0.833&0.667&0.752&0.650&0.733&0.708&0.611&&0.48&0.17&0.59&0.83&0.60&0.69&0.57&0.67&0.64&0.48\\
&SVM &0.944&0.917&0.944&0.278&0.639&0.944&0.944&0.917&0.944&0.944&& 0.94&0.91&0.94&0.12&0.52&0.94&0.94&0.91&0.94&0.94\\
&KNN &0.917&0.889&0.917&0.583&0.639&0.917&0.889&0.861&0.917&0.917&& 0.91&0.88&0.91&0.47&0.52&0.91&0.88&0.85&0.91&0.91\\
&DTC &0.694&0.694&0.778&0.639&0.778&0.750&0.694&0.667&0.639&0.805&& 0.70&0.68&0.77&0.52&0.76&0.75&0.70&0.65&0.64&0.80\\
&RFC &0.889&0.805&0.861&0.861&0.889&0.861&0.944&0.889&0.833&0.889&& 0.89&0.79&0.85&0.86&0.89&0.86&0.94&0.89&0.83&0.89\\
&SCL-TPE&\textbf{0.972} & & & & & & & & & &&\textbf{0.97}\\
\specialrule{0em}{0.6pt}{0.6pt}
\hline
\specialrule{0em}{0.6pt}{0.6pt}
\multirow{6}{*}{lym} & MLP &0.806&0.484&0.710&0.613&0.677&0.839&/&/&0.838&0.871&& 0.79&0.39&0.69&0.60&0.62&0.82&/&/&0.83&0.86\\
&SVM &0.774&0.516&0.742&0.516&0.516&0.839&/&/&0.709&0.806&& 0.75&0.51&0.72&0.35&0.35&0.84&/&/&0.70&0.79\\
&KNN &0.742& / &0.710&0.452&0.581&0.742&/&/&0.677&0.774&& 0.72& / &0.69&0.44&0.51&0.73&/&/&0.67&0.76 \\
&DTC &0.645&0.452&0.742&0.484&0.323&0.613&/&/&0.645&0.710&& 0.64&0.43& 0.71&0.41&0.33&0.61&/&/&0.63&0.70\\
&RFC &0.871&0.613&0.742&0.548&0.548&0.806&/&/&0.903&0.871&& 0.84&0.64& 0.71&0.53&0.40&0.83&/&/&0.89&0.86\\
&SCL-TPE&\textbf{0.903} & & & & & & & & & &&\textbf{0.90}\\
\specialrule{0em}{0.6pt}{0.6pt}
\hline
\specialrule{0em}{0.6pt}{0.6pt}

\multirow{6}{*}{gla} & MLP &0.523&0.231&0.431&0.539&0.415&0.523&0.477&/&0.446&0.585&& 0.47&0.18&0.33&0.49&0.40&0.49&0.43&/&0.36&0.55\\
&SVM &0.369&0.292&0.354&0.369&0.262&0.308&0.385&/&0.400&0.400&&0.27&0.29&0.21&0.26&0.15&0.25&0.30&/&0.31&0.27\\
&KNN &0.569&0.523&0.538&0.277&0.400&0.538&0.492&/&0.523&0.538&&0.55&0.52&0.50&0.16&0.40&0.55&0.49&/&0.53&0.54\\
&DTC &0.554&0.338&0.477&0.462&0.400&0.508&0.431&/&0.446&0.538&&0.47&0.26&0.39&0.41&0.33&0.51&0.42&/&0.49&0.42\\
&RFC&0.600&0.492&0.538&0.462&0.354&0.600&0.585&/&0.554&0.554&&0.57&0.46&0.48&0.42&0.29&0.57&0.56&/&0.52&0.52\\
&SCL-TPE&\textbf{0.646} & & & & & & & & & &&\textbf{0.61}\\
\specialrule{0em}{0.6pt}{0.6pt}
\hline
\specialrule{0em}{0.6pt}{0.6pt}

\multirow{6}{*}{page} & MLP &0.915&0.867&0.933&0.909&0.036&0.945&0.939&/&0.952&0.927&&0.92&0.86&0.91&0.87&0.00&0.89&0.93&/&0.95&0.93\\
&SVM &0.909&0.673&0.903&0.897&0.024&0.794&0.776&/&0.769&0.939&&0.87&0.77&0.86&0.85&0.00&0.85&0.84&/&0.83&0.94\\
&KNN &0.945&0.933&0.921&0.903&0.909&0.915&0.879&/&0.909&0.952&&0.93&0.92&0.90&0.86&0.87&0.92&0.89&/&0.92&0.94\\
&DTC &0.927&0.758&0.933&0.406&0.582&0.903&0.909&/&0.903&0.921&&0.93&0.80&0.91&0.53&0.66&0.90&0.91&/&0.90&0.93\\
&RFC &0.952&0.739&0.939&0.903&0.618&0.933&0.933&/&0.939&0.927&&0.95&0.81&0.92&0.89&0.71&0.92&0.92&/&0.93&0.93\\
&SCL-TPE&\textbf{0.964} & & & & & & & & & &&\textbf{0.96}\\
\specialrule{0em}{0.6pt}{0.6pt}
\hline
\specialrule{0em}{0.6pt}{0.6pt}

\multirow{6}{*}{dt} & MLP &0.819&0.833&0.944&0.514&0.681&0.833&0.958&0.972&0.792&0.972 &&0.75&0.76&0.95&0.41&0.57&0.78&0.96&0.97&0.72&0.97 \\
&SVM &\textbf{0.986}&0.944&0.944&0.056&0.542&0.958&0.972&0.986&0.972&0.931 &&\textbf{0.99}&0.94&0.97&0.01&0.45&0.96&0.97&\textbf{0.99}&0.97&0.93\\
&KNN &0.958&0.931&0.958&0.569&0.389&0.958&0.931&0.958&0.944&0.958&&0.96&0.93&0.96&0.51&0.24&0.96&0.93&0.96&0.94&0.96\\
&DTC &0.847&0.889&0.958&0.625&0.625&0.903&0.903&0.931& 0.958&0.875&& 0.84&0.88&0.96&0.62&0.63&0.90&0.91&0.93&0.96&0.87 \\
&RFC&0.972&0.972&0.972&0.653&0.736&0.958&0.958&0.958&0.958&0.972&& 0.97&0.97&0.97&0.63&0.66&0.96&0.96&0.96&0.96 &0.97\\
&SCL-TPE&\textbf{0.986} & & & & & & & &  & &&\textbf{0.99}\\
\specialrule{0em}{0.6pt}{0.6pt}
\hline
\specialrule{0em}{0.6pt}{0.6pt}

\multirow{6}{*}{pb} & MLP &0.782&0.673&0.677&0.591&0.691&0.741&0.777&0.791&0.745&0.873 &&0.73&0.60&0.60&0.51&0.62&0.68&0.72&0.74&0.70&0.84 \\
&SVM &0.905&0.905&0.905&0.095&0.414&0.905&0.914&0.905&0.909&0.900 &&0.90&0.90&0.90&0.02&0.36&0.90&0.91&0.90&0.91&0.90 \\
&KNN &0.941&0.941&0.932&0.473&0.586&0.932&0.950&0.941&0.945&0.941 &&0.94&0.94&0.93&0.44&0.52&0.93&0.95&0.94&0.94&0.94 \\
&DTC &0.900&0.918&0.909&0.541&0.718&0.868&0.877&0.855&0.845&0.895 &&0.90&0.92&0.91&0.53&0.71&0.87&0.88&0.85&0.84&0.89\\
&RFC &0.955&0.959&0.941&0.664&0.705&0.959&0.959&0.955&0.959&0.964 &&0.95&0.96&0.94&0.62&0.66&0.96&0.96&0.95&0.96&0.96\\
&SCL-TPE&\textbf{0.968} & & & & & & & &  & &&\textbf{0.97}\\
\bottomrule
\end{tabular}
\end{table}

\begin{table}[!h]
\tiny
\centering
\tabcolsep=0.05cm
\caption{\centering G-mean and AUC score for multi-class imbalanced data.}\label{multi2}
\begin{tabular}{cccccccccccccccccccccccc}
\toprule
\multirow{2}{*}{Data}& &\multicolumn{10}{c}{G-mean}&\makebox[0.01\textwidth][c]{}&\multicolumn{10}{c}{AUC}\\
\cline{3-12}
\cline{14-23}
\rule{0pt}{11pt}
&  & None&RUS&ENN&CNN&OSS&ROS&SMO&ADA&BSM&GAN&\makebox[0.01\textwidth][c]{}&\makebox[0.037\textwidth][c]{None}&\makebox[0.037\textwidth][c]{RUS}&\makebox[0.035\textwidth][c]{ENN}&\makebox[0.04\textwidth][c]{CNN}&\makebox[0.037\textwidth][c]{OSS}&\makebox[0.037\textwidth][c]{ROS}&\makebox[0.037\textwidth][c]{SMO}&\makebox[0.037\textwidth][c]{ADA}&\makebox[0.037\textwidth][c]{BSM}&\makebox[0.037\textwidth][c]{GAN}\\

\midrule
\multirow{6}{*}{bal} & MLP &0.94&0.96&0.90&0.98&0.94&0.99&0.98&0.93&0.99&0.96&&0.866&0.961&0.845&0.943&0.866&0.976&0.943&0.884&0.981&0.904\\
&SVM &0.85&0.84&0.86&0.88&0.84&0.93&0.94&0.94&0.94&0.85&& 0.786&0.852&0.792&0.808&0.781&0.935&0.939&0.943&0.939&0.787\\
&KNN &0.79&0.85&0.84&0.86&0.80&0.78&0.79&0.80&0.80&0.80&& 0.733&0.864&0.817&0.815&0.738&0.749&0.758&0.767&0.771&0.739\\
&DTC &0.84&0.79&0.83&0.74&0.84&0.74&0.83&0.80&0.83&0.84&& 0.814&0.783&0.783&0.716&0.809&0.716&0.779&0.782&0.787&0.814\\
&RFC &0.78&0.82&0.80&0.82&0.77&0.73&0.80&0.79&0.78&0.80&& 0.743&0.801&0.760&0.774&0.736&0.708&0.767&0.762&0.743&0.767\\
&SCL-TPE&\textbf{1.0} & & & & & & & & & &&\textbf{1.0}\\
\specialrule{0em}{0.6pt}{0.6pt}
\hline
\specialrule{0em}{0.6pt}{0.6pt}
\multirow{6}{*}{wine} & MLP &0.40&0.00&0.62&0.87&0.68&0.68&0.53&0.63&0.63&0.50&& 0.668&0.500&0.743&0.885&0.752&0.828&0.739&0.805&0.778&0.737\\
&SVM &0.96&0.94&0.96&0.00&0.53&0.96&0.96&0.94&0.96&0.96&&0.963&0.944&0.963&0.500&0.724&0.963&0.963&0.944&0.963&0.963\\
&KNN &0.94&0.91&0.94&0.50&0.56&0.94&0.91&0.89&0.94&0.94&&0.944&0.925&0.944&0.690&0.733&0.944&0.925&0.907&0.944&0.944\\
&DTC &0.76&0.76&0.83&0.53&0.82&0.81&0.76&0.74&0.71&0.85&&0.771&0.781&0.840&0.724&0.847&0.813&0.771&0.757&0.725&0.861\\
&RFC &0.92&0.84&0.89&0.89&0.91&0.89&0.96&0.91&0.87&0.91&&0.924&0.870&0.906&0.902&0.917&0.902&0.959&0.923&0.885&0.923\\
&SCL-TPE&\textbf{0.98} & & & & & & & & & &&\textbf{0.981}\\
\specialrule{0em}{0.6pt}{0.6pt}
\hline
\specialrule{0em}{0.6pt}{0.6pt}
\multirow{6}{*}{lym} & MLP &0.81&0.37&0.70&0.62&0.58&0.82&/&/&0.83&0.86&&0.788&0.644&0.621&0.688&0.591&0.798&/&/&0.802&0.817\\
&SVM &0.73&0.58&0.70&0.00&0.00&0.84&/&/&0.72&0.79&&0.647&0.771&0.632&0.500&0.500&0.917&/&/&0.740&0.783\\
&KNN &0.72&/&0.68&0.44&0.42&0.74&/&/&0.68&0.77&&0.635&/&0.617&0.490&0.539&0.755&/&/&0.726&0.769\\
&DTC &0.65&0.50&0.71&0.38&0.33&0.63&/&/&0.65&0.73&& 0.601&0.630&0.633&0.526&0.566&0.693&/&/&0.711&0.750\\
&RFC&0.85&0.70&0.71&0.54&0.26&0.85&/&/&0.89&0.87&& 0.703&0.712&0.633&0.538&0.639&0.912&/&/&0.835&0.820\\
&SCL-TPE&\textbf{0.90} & & & & & & & & & &&\textbf{0.949}\\
\specialrule{0em}{0.6pt}{0.6pt}
\hline
\specialrule{0em}{0.6pt}{0.6pt}

\multirow{6}{*}{gla} & MLP &0.51&0.26&0.37&0.57&0.49&0.61&0.52&/&0.40&0.64&&0.640&0.535&0.600&0.718&0.702&0.727&0.652&/&0.579&0.726\\
&SVM &0.34&0.44&0.24&0.26&0.16&0.34&0.46&/&0.43&0.33&&0.585&0.568&0.588&0.577&0.510&0.674&0.743&/&0.753&0.710\\
&KNN &0.62&0.64&0.58&0.26&0.57&0.67&0.66&/&0.65&0.66&&0.741&0.717&0.663&0.610&0.697&0.746&0.712&/&0.734&0.729\\
&DTC &0.54&0.39&0.45&0.56&0.46&0.64&0.65&/&0.62&0.54&&0.728&0.725&0.648&0.757&0.713&0.706&0.705&/&0.726&0.687\\
&RFC &0.64&0.56&0.54&0.55&0.41&0.65&0.57&/&0.60&0.64&&0.745&0.755&0.680&0.720&0.741&0.751&0.744&/&0.726&0.764\\
&SCL-TPE&\textbf{0.69} & & & & & & & & & &&\textbf{0.787}\\
\specialrule{0em}{0.6pt}{0.6pt}
\hline
\specialrule{0em}{0.6pt}{0.6pt}

\multirow{6}{*}{page} & MLP &0.79&0.56&0.67&0.50&0.02&0.61&0.79&/&0.89&0.93&& 0.829&0.652&0.645&0.727&0.601&0.741&0.705&/&0.807&0.846\\
&SVM &0.33&0.77&0.24&0.22&0.02&0.84&0.83&/&0.82&0.86&&0.532&0.815&0.516&0.505&0.501&0.809&0.805&/&0.769&0.864\\
&KNN &0.72&0.77&0.57&0.31&0.33&0.84&0.85&/&0.87&0.76&&0.817&0.721&0.614&0.611&0.637&0.790&0.812&/&0.819&0.848\\
&DTC &0.88&0.70&0.67&0.47&0.44&0.77&0.77&/&0.76&0.90&&0.841&0.709&0.776&0.645&0.637&0.823&0.824&/&0.723&0.806\\
&RFC &0.86&0.81&0.67&0.70&0.61&0.75&0.75&/&0.78&0.85&&0.793&0.848&0.786&0.829&0.572&0.814&0.814&/&0.830&0.852\\
&SCL-TPE&\textbf{0.95} & & & & & & & & & &&\textbf{0.869}\\
\specialrule{0em}{0.6pt}{0.6pt}
\hline
\specialrule{0em}{0.6pt}{0.6pt}

\multirow{6}{*}{dt} & MLP &0.79&0.80&0.97&0.54&0.62&0.82&0.97&0.98&0.75&0.98 &&0.886&0.897&0.972&0.695&0.787&0.875&0.974&0.985&0.888&0.982\\
&SVM &\textbf{0.99}&0.96&0.98&0.00&0.49&0.97&0.98&\textbf{0.99}&0.98&0.95 &&0.990&0.967&0.981&0.500&0.704&0.975&0.983&0.990&0.983&0.959\\
&KNN &0.98&0.96&0.98&0.63&0.29&0.98&0.96&0.98&0.97&0.98&&0.978&0.965&0.978&0.771&0.607&0.977&0.964&0.978&0.967&0.978\\
&DTC &0.90&0.92&0.97&0.74&0.73&0.93&0.94&0.96&0.98&0.92&&0.873&0.935&0.974&0.747&0.747&0.938&0.924&0.957&0.975&0.918\\
&RFC&0.98&0.98&0.98&0.74&0.74&0.97&0.97&0.97&0.97&0.98&& 0.982&0.982&0.982&0.787&0.827&0.973&0.973&0.973&0.973&0.982\\
&SCL-TPE&\textbf{0.99} & & & & & & & &  & &&\textbf{0.991}\\
\specialrule{0em}{0.6pt}{0.6pt}
\hline
\specialrule{0em}{0.6pt}{0.6pt}

\multirow{6}{*}{pb} & MLP &0.78&0.66&0.67&0.57&0.68&0.75&0.77&0.79&0.76&0.88&& 0.874&0.820&0.819&0.772&0.822&0.855&0.876&0.879&0.858&0.927\\
&SVM &0.94&0.94&0.94&0.01&0.46&0.94&0.95&0.94&0.95&0.94 &&0.947&0.947&0.944&0.500&0.674&0.947&0.952&0.947&0.950 & 0.944\\
&KNN &0.97&0.97&0.96&0.60&0.67&0.96&0.97&0.97&0.97&0.97 &&0.967&0.967&0.962&0.706&0.771&0.963&0.973&0.967&0.970& 0.967\\
&DTC &0.94&0.95&0.95&0.69&0.82&0.92&0.93&0.91&0.91&0.94 &&0.945&0.954&0.949&0.746&0.845&0.927&0.933&0.920&0.915&0.943 \\
&RFC &0.97&\textbf{0.98}&0.97&0.76&0.76&\textbf{0.98}&0.97&\textbf{0.98}&\textbf{0.98}&\textbf{0.98}&&0.975&0.978&0.967&0.815&0.838&0.978&0.975&0.978&0.978&0.981 \\
&SCL-TPE&\textbf{0.98} & & & & & & & &  & &&\textbf{0.983}\\
\bottomrule
\end{tabular}
\end{table}
The proposed method is also compared with ten sampling methods tested on five base classifiers in the multi-class classification task. From Tables \ref{multi1} and \ref{multi2}, we can see the proposed method outperforms other approaches on all four metrics. For example, on the Wine dataset, the proposed method yields an accuracy result of 0.972, which is 2.8$\%$ better than the second best method (GAN with SVM); F-measure of 0.97, which is 3$\%$ higher than the second best method; G-mean of 0.98, which is 2$\%$ higher than the second-best method; and AUC of 0.981, which is 1.8$\%$ superior to the second best method. It is worth noting that there are only one or two samples of the minority class in the test datasets of the lymphography, glass, and pageblocks, resulting in the failure of SMOTE and ADASYN. The excellent performance of SCL-TPE in these datasets also proves the robustness of our method.

\subsubsection{Ablation study}
\begin{table}[!h] 
\tiny
\setlength{\tabcolsep}{2.7pt}
\centering
\caption{Ablation study on fifteen datasets.}
\begin{tabular}{|c|c|c|c|c|c|c|c|c|c|c|c|c|c|c|c|c|}
\hline{Metrics} &{Methods} &{gl0} &{eo2} &{yt3} &{yt6}&{vw0}&{hb}&{yt24} &{pa0}&{bal}&{wine}&{lym}&{gla}&{page}&{dt}&{pb}\\ \hline
\multirow{2}{*}{Accuracy}
&SCL&0.767&0.794&0.899&0.922&0.985&0.629&0.883&0.890&0.960&\textbf{0.971}&0.774&0.185&0.703&0.958&0.527 \\
&CL-TPE&0.744&0.735&0.852&0.892&0.874&0.581&0.864 &0.853&0.864&0.944&0.677&0.415&0.927&0.861&0.786 \\
&SCL-TPE & \textbf{0.837}&\textbf{0.956}&\textbf{0.919}&\textbf{0.976}&\textbf{1.0}&\textbf{0.645}&\textbf{0.971}&\textbf{0.918}&\textbf{1.0}&\textbf{0.971}&\textbf{0.903}&\textbf{0.646}&\textbf{0.964}&\textbf{0.986}& \textbf{0.968}\\
\hline
\multirow{2}{*}{F-measure}
&SCL&0.77&0.81&0.91&0.94&0.99&0.65&0.90&0.91 &0.96&\textbf{0.97}&0.77&0.19&0.77&0.96&0.51\\
&CL-TPE&0.75&0.77&0.87&0.93&0.89&0.60&0.89&0.88&0.89&0.94&0.68&0.38&0.93&0.86&0.78\\
&SCL-TPE &\textbf{0.84}&\textbf{0.96}&\textbf{0.93}&\textbf{0.98}&\textbf{1.0}&\textbf{0.66}&\textbf{0.97}&\textbf{0.93}&\textbf{1.0}&\textbf{0.97}&\textbf{0.90}&\textbf{0.61}&\textbf{0.96}&\textbf{0.99}&\textbf{0.97}\\
\hline
\multirow{2}{*}{G-mean} 
&SCL&0.73&0.77&0.88&0.89&0.99&0.66&0.93&0.92&0.98&\textbf{0.98}&0.79&0.29&0.76&0.97&0.59\\
&CL-TPE&0.77&0.83&0.89&\textbf{0.94}&0.93&0.61&0.92&0.90&0.92 &0.96&0.70&0.51&0.90&0.91&0.87\\
&SCL-TPE &\textbf{0.84}&\textbf{0.90}&\textbf{0.95}&0.92&\textbf{1.0}&\textbf{0.68}&\textbf{0.98}&\textbf{0.94}&\textbf{1.0}&\textbf{0.98}&\textbf{0.90}&\textbf{0.69}&\textbf{0.95}&\textbf{0.99}&\textbf{0.98}\\
\hline
\multirow{2}{*}{AUC}
&SCL&0.735&0.767&0.877&0.891&0.992&0.671&0.935&0.919&0.978&\textbf{0.981}&0.776&0.593&0.783& 0.972&0.739\\
&CL-TPE&0.773&0.842&0.890&\textbf{0.944}&0.931&0.620&0.925 &0.902&0.911&0.957&0.840&0.706&0.837&0.914&0.881\\
&SCL-TPE &\textbf{0.842}&\textbf{0.900}&\textbf{0.955}&0.918&\textbf{1.0}&\textbf{0.682}&\textbf{0.984}&\textbf{0.942}&\textbf{1.0}&\textbf{0.981}&\textbf{0.949}&\textbf{0.787}&\textbf{0.869}&\textbf{0.991}&\textbf{0.983}\\
\hline
\end{tabular}
\label{ablation}
\end{table}

\begin{figure}[!h]
\centering
\subfigure[yeast3]{
\includegraphics[width=5.5cm]{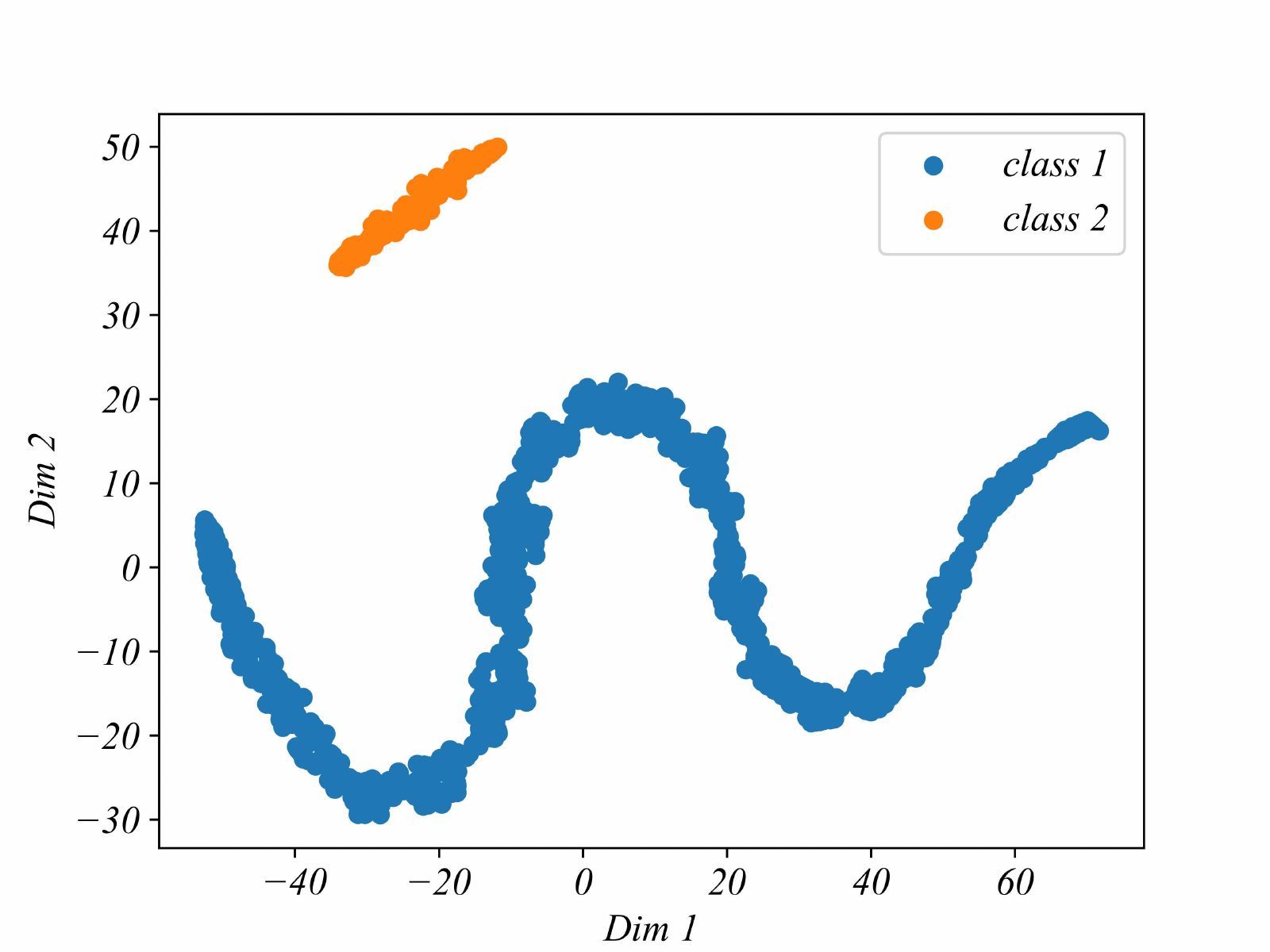}
}
\quad
\subfigure[vowel0]{
\includegraphics[width=5.5cm]{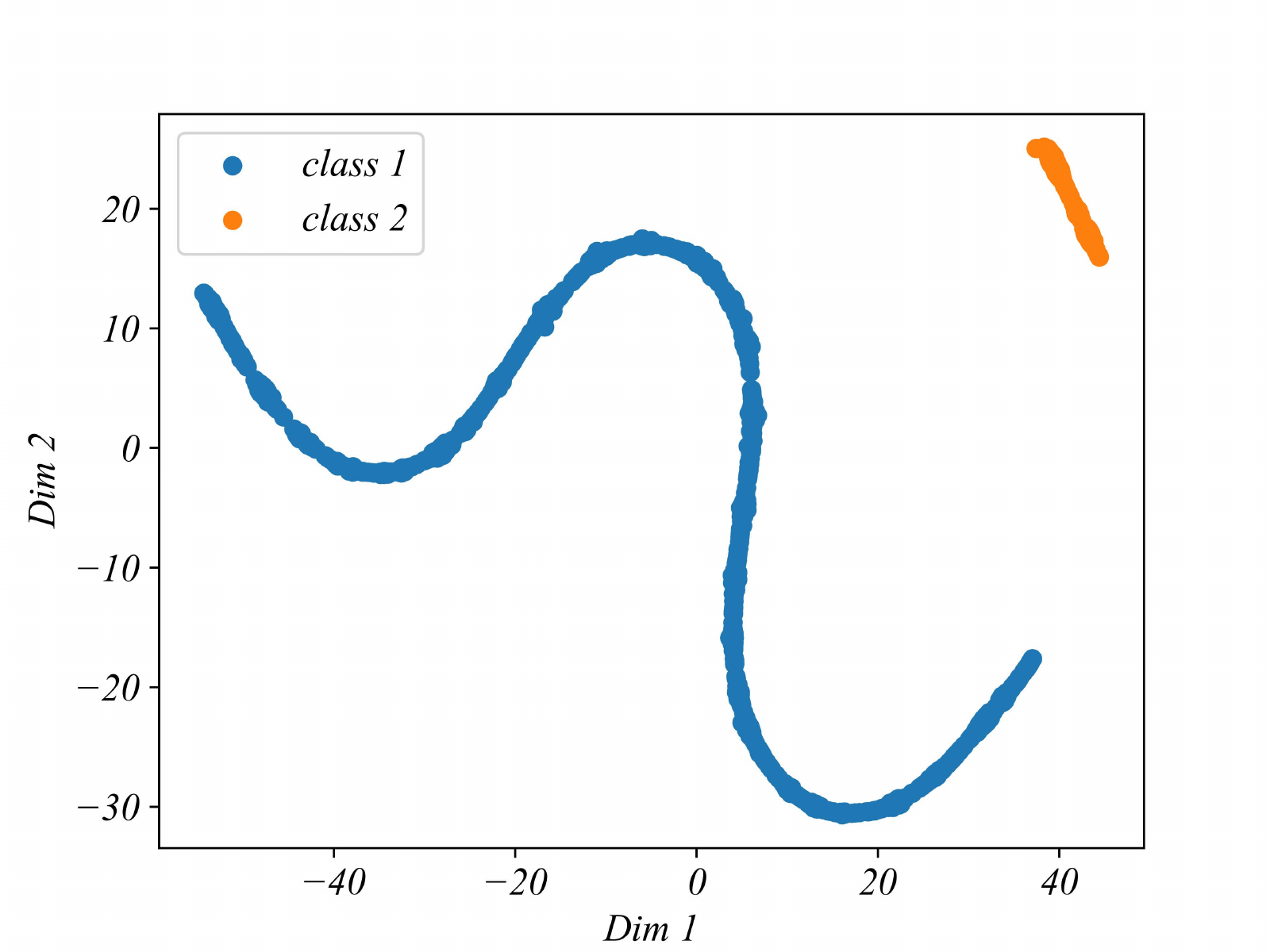}
}
\quad
\subfigure[lymphography]{
\includegraphics[width=5.5cm]{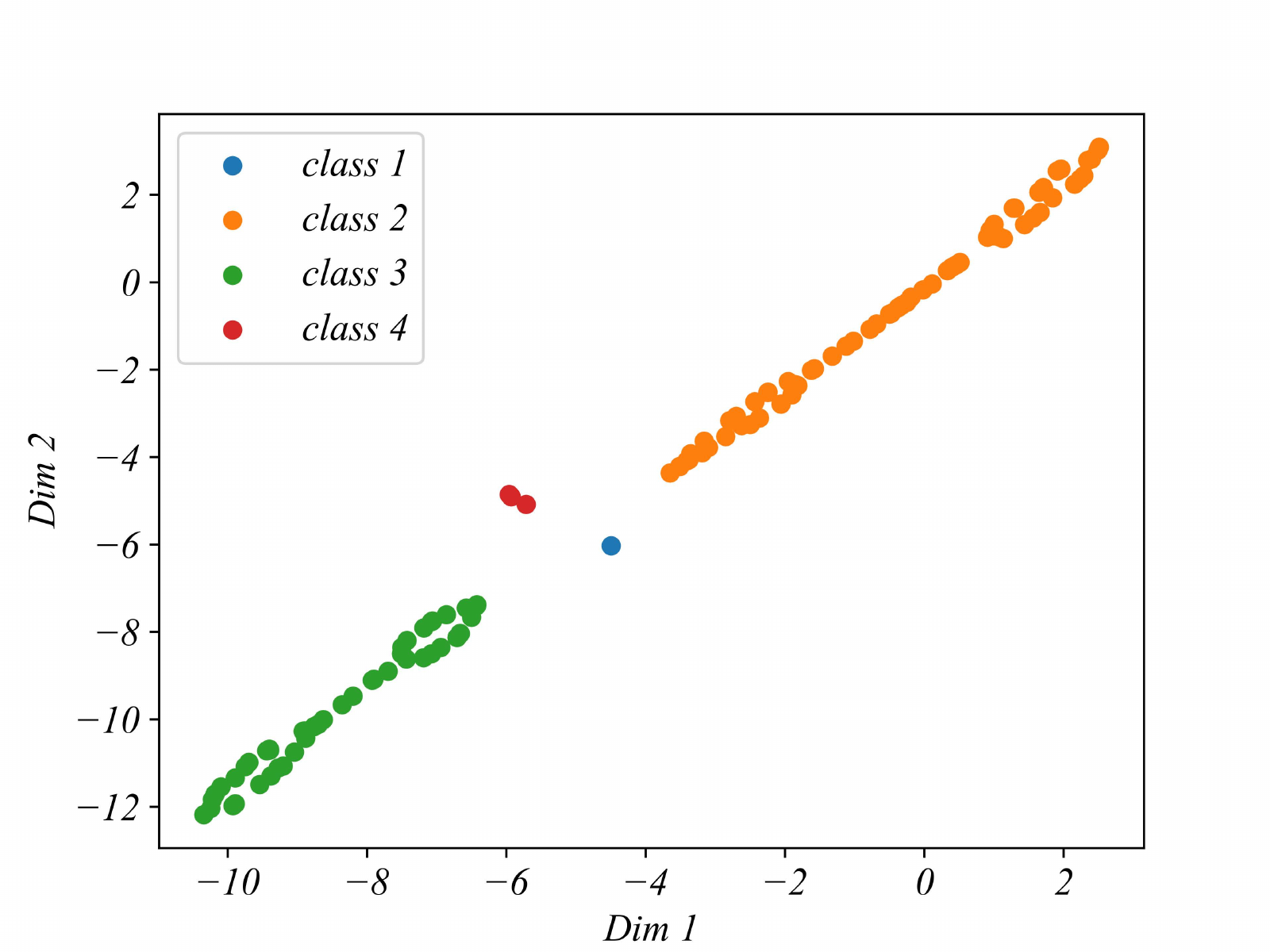}
}
\quad
\subfigure[glass]{
\includegraphics[width=5.5cm]{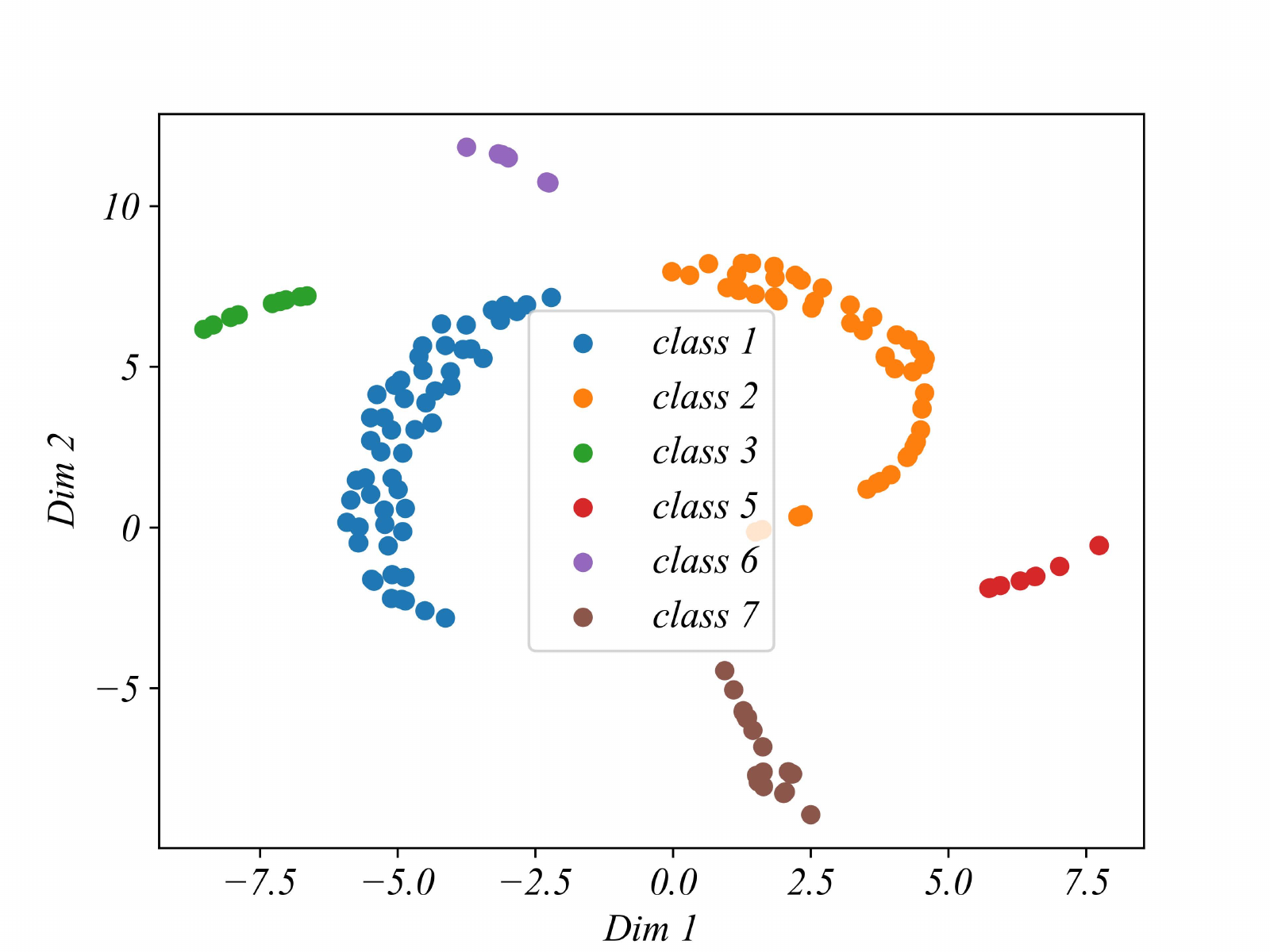}
}
\caption{The t-SNE embeddings of yeast3, vowel0, lymphography, glass.}\label{tsne}
\end{figure}

In addition to the advanced resampling methods, we conduct an ablation study to analyze the performance gain of each component in SCL-TPE on the imbalanced tabular datasets. We define two variants of SCL-TPE: 1. SCL only. We exclude the TPE and fix temperature $\tau$ as 0.5. 2. CL-TPE. We exclude the supervised contrastive loss and use contrastive loss instead. 

Ablations are provided in Table \ref{ablation}. From experimental results, SCL-TPE always performs better than its variants. We conclude that each of the major elements of our method is valuable and crucial for performance, and the best performance is achieved when they work collaboratively.

\begin{figure}[!h]
\centering
\subfigure[yeast3]{
\includegraphics[width=2.3cm]{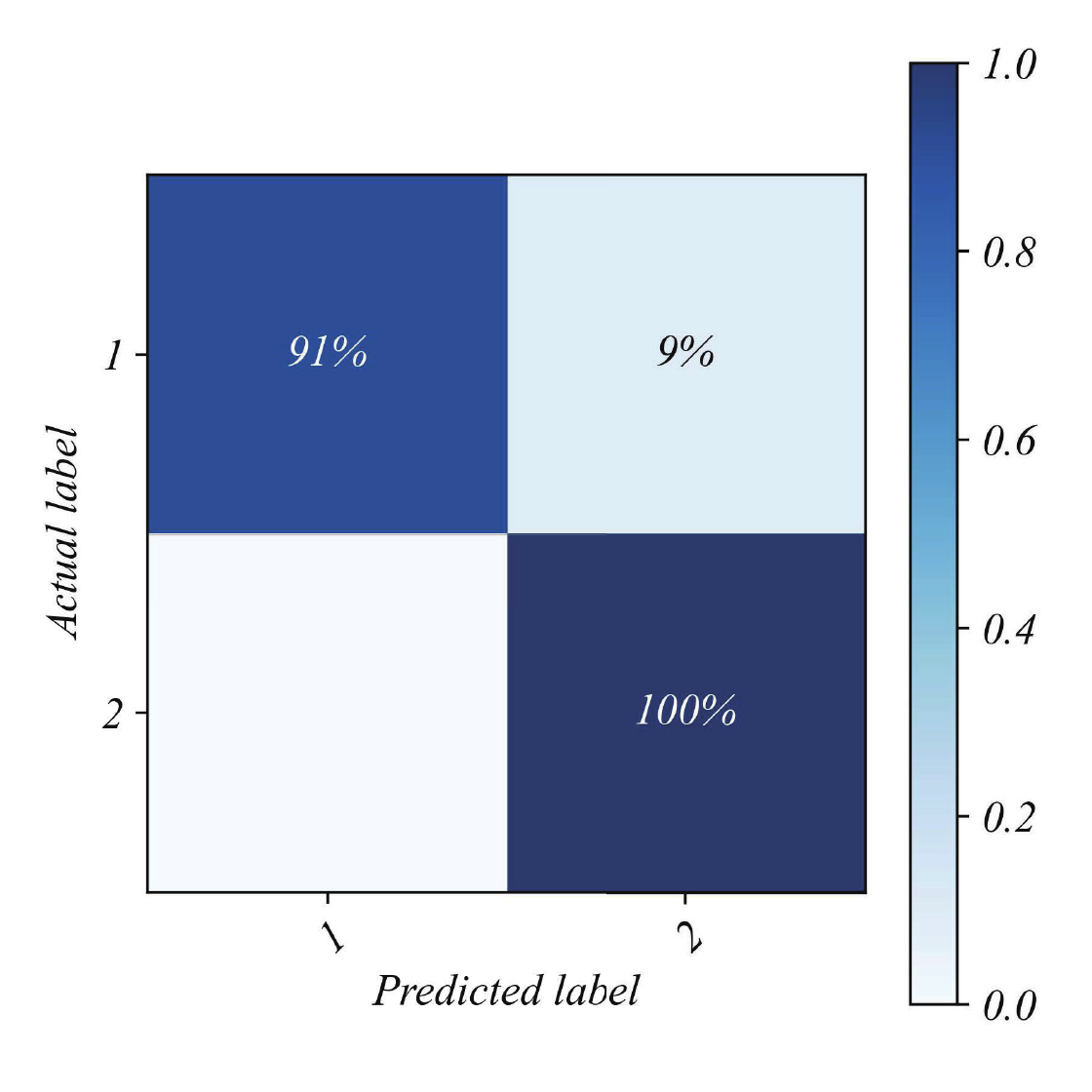}
}
\quad
\subfigure[vowel0]{
\includegraphics[width=2.3cm]{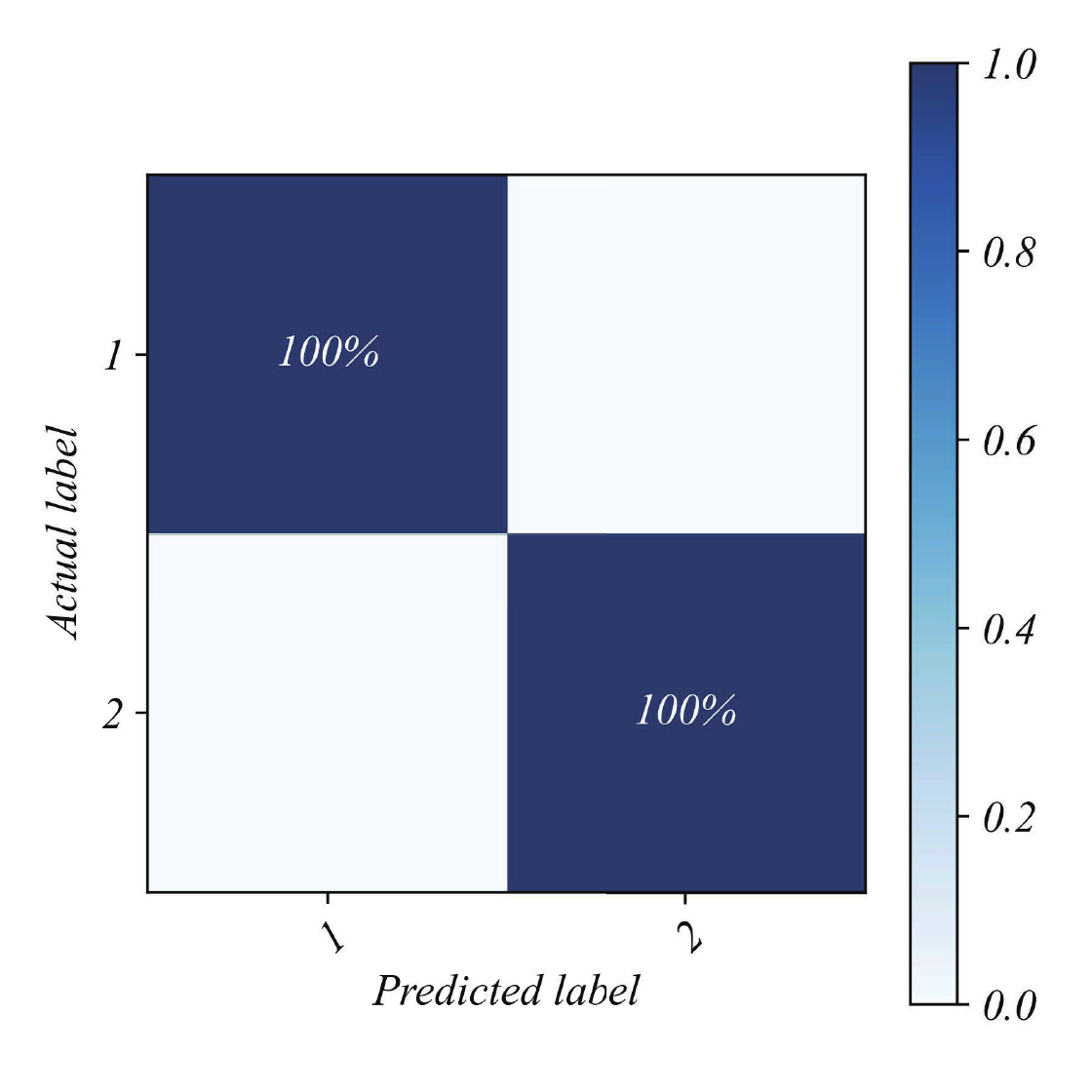}
}
\centering
\quad
\subfigure[lymphography]{
\includegraphics[width=3.1cm]{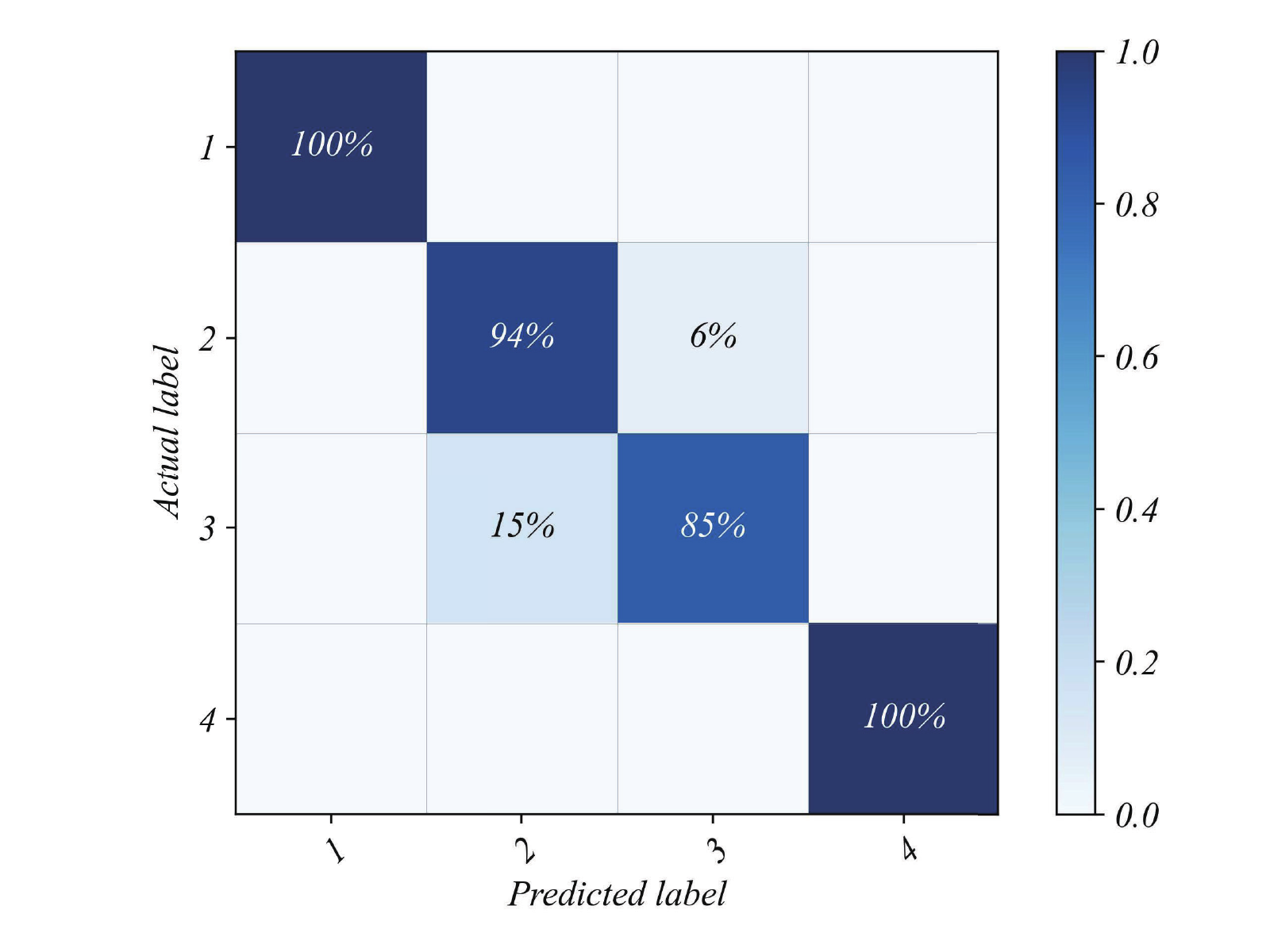}
}
\quad
\subfigure[glass]{
\includegraphics[width=3.1cm]{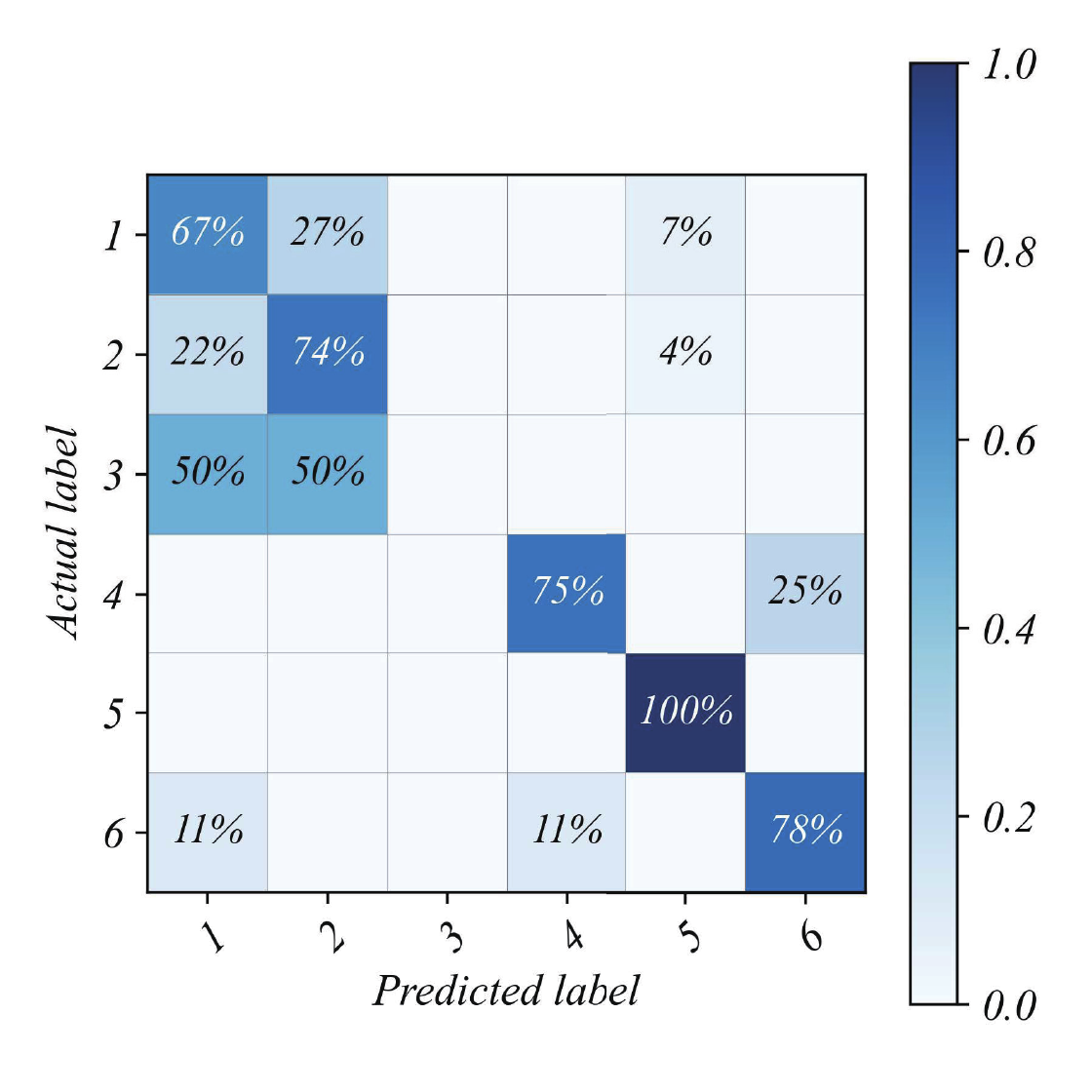}
}
\caption{The confusion matrices of yeast3, vowel0, lymphography, glass.}\label{cm}
\end{figure}

Finally, the t-SNE \cite{van2008visualizing} embeddings of two binary datasets and two multiclass datasets are shown in Fig. \ref{tsne}. The discriminative and compact representation learned by SCL is very helpful for downstream imbalanced dataset classification. 

To ensure that the final performance of SCL-TPE is effective, we report the confusion matrices of the proposed method on these four datasets in Fig. \ref{cm}, where the rows and columns represent prediction classes and actual classes, respectively. The color of each grid represents a specific value, which means the ratio of the number of correctly predicted samples to the total number of actual samples. The darker the color of diagonal blocks and the lighter the color of other blocks, the better the effect of the model.

\section{Conclusions and future work}
\label{sec:conclusion}
In this study, we propose a novel SCL-TPE for imbalanced tabular data. In representation learning, SCL learns the pattern information hidden in the data based on contrastive loss and further leverages the label information, which addresses the limited data augmentation techniques of tabular data. We reveal the significant influence of the hyper-parameter $\tau$ on the model performance and introduce TPE to select the best $\tau$ automatically. We demonstrate that TPE surpasses three other HPO methods: grid search, random search, and genetic algorithm. Experimental results on binary and multi-class datasets prove that the proposed method obtains better performance than other methods in terms of four metrics: Accuracy, F-measure, G-mean, and AUC.

There are still several works to do in the future. First, it is of interest to investigate the issue of inadequate data augmentation techniques for tabular data. More data augmentation techniques \cite{shenkar2021anomaly} for tabular data can be proposed, so that the performance of CL, which shows potential for imbalanced image classification, can be further enhanced in the tabular data field. Second, we will continue to study other HPO methods for selecting temperature $\tau$. A comprehensive comparison can be carried out based on the intention to combining SCL with other newly proposed HPO methods such as Heteroscedastic and Evolutionary Bayesian Optimisation solver (HEBO) \cite{cowen2020empirical}.




\section{Acknowledgments}
This work was supported by the National Key R\&D Program of China under Grant No. 2020YFB1707803. This work was also supported in part by the Zhejiang University/University of Illinois at Urbana-Champaign Institute, and was led by Principal Supervisor Prof. Hongwei Wang.
\clearpage
\bibliographystyle{model1-num-names}
\bibliography{SCL-TPE.bib}







\clearpage

\end{document}